%% file: main.tex
\pdfoutput=1
\documentclass[11pt]{article}

\usepackage[]{acl}

\usepackage{arabtex}
\usepackage{utf8}
\setcode{utf8}
\usepackage{subcaption}

\usepackage[T1]{fontenc}
\usepackage[utf8]{inputenc}

\usepackage{microtype}

\usepackage{inconsolata}
\usepackage{microtype}
\usepackage{graphicx}
\usepackage{import}
\usepackage{layout}
\usepackage{tabularx, makecell}
\usepackage{booktabs}
\usepackage{mathrsfs}
\usepackage{amssymb} 
\usepackage{url}
\usepackage{graphicx}
\usepackage{xspace,paralist}
\usepackage{times,latexsym}
\usepackage{amsmath}
\usepackage{appendix}
\usepackage{comment} 
\usepackage{enumitem}
\usepackage{makecell}
\usepackage{multirow}
\usepackage{xcolor}
\usepackage{cleveref}
\usepackage{todonotes}
\usepackage{longtable,supertabular}
\usepackage[T2A,T1]{fontenc}
\usepackage[russian,english]{babel}
\usepackage{array}
\usepackage{CJKutf8}
\usepackage{amssymb}% http://ctan.org/pkg/amssymb
\usepackage{pifont}% http://ctan.org/pkg/pifont

\newcommand{\tabref}[2][]{Table#1~\ref{#2}\xspace}
\newcommand{\figref}[1]{Figure~\ref{#1}\xspace}

\newcommand{\appref}[1]{Appendix~\ref{#1}\xspace}

\newcommand{\dataset}[1]{\text{#1}\xspace}
\newcommand{\arxiv}{\dataset{arXiv}}
\newcommand{\wikipedia}{\dataset{Wikipedia}}
\newcommand{\wikihow}{\dataset{WikiHow}}
\newcommand{\peerread}{\dataset{PeerRead}}
\newcommand{\reddit}{\dataset{Reddit}}
\newcommand{\outfox}{\dataset{OUTFOX}}

\newcommand{\model}[1]{\text{#1}\xspace}

\newcommand{\chatgpt}{\model{ChatGPT}}
\newcommand{\gptfour}{\model{GPT-4}}

\newcommand{\davinci}{\model{davinci-003}}

\newcommand{\cohere}{\model{Cohere}}
\newcommand{\dolly}{\model{Dolly-v2}}
\newcommand{\bloomz}{\model{BLOOMz}}

\newcommand{\llamatwo}{\model{LLaMA-2}}

\newcommand{\roberta}{\model{RoBERTa}}
\newcommand{\xlmr}{\model{XLM-R}}
\newcommand{\gltr}{\model{GLTR}}
\newcommand{\nela}{\model{NELA}}

\title{M4GT-Bench: Evaluation Benchmark\\ for Black-Box Machine-Generated Text Detection}

\author{
Yuxia Wang,\textsuperscript{\textdagger} 
Jonibek Mansurov,\textsuperscript{\textdagger} 
Petar Ivanov,\textsuperscript{\textdagger}
Jinyan Su,\textsuperscript{\textdagger} \\ 
{\bf Artem Shelmanov,\textsuperscript{\textdagger}
Akim Tsvigun,\textsuperscript{\textdagger}
Osama Mohammed Afzal,\textsuperscript{\textdagger}
}\\
{\bf Tarek Mahmoud,\textsuperscript{\textdagger} 
Giovanni Puccetti,\textsuperscript{\S}
Thomas Arnold,\textsuperscript{\textparagraph}
}\\
{\bf Alham Fikri Aji,\textsuperscript{\textdagger} 
Nizar Habash,\textsuperscript{\textdagger}\textsuperscript{\textdaggerdbl} 
Iryna Gurevych,\textsuperscript{\textdagger} 
Preslav Nakov\textsuperscript{\textdagger}
}\\
  \textsuperscript{\textdagger}Mohamed bin Zayed University of Artificial Intelligence, UAE \\ 
  \textsuperscript{\textparagraph}TU Darmstadt, Germany 
  \textsuperscript{\S}Institute of Information Science and Technology, Italy\\  
  \textsuperscript{\textdaggerdbl}New York University Abu Dhabi, UAE\\
  \texttt{\{yuxia.wang, jonibek.mansurov, preslav.nakov\}@mbzuai.ac.ae} \\
}
\begin{document}
\maketitle
\begin{abstract}
% The advent of Large Language Models (LLMs) has brought
% a shift in the landscape of textual content generation. 
% LLMs such as ChatGPT and GPT-4 have become pervasive and easily accessible. Their remarkable fluency and versatility in text generation have catalyzed an 
The advent of Large Language Models (LLMs) has brought an unprecedented surge in machine-generated text (MGT) across diverse channels. This raises legitimate concerns about its potential misuse and societal implications. The need to identify and differentiate such content from genuine human-generated text is critical in combating disinformation, preserving the integrity of education and scientific fields, and maintaining trust in communication. In this work, we address this problem by introducing a new benchmark based on a \textbf{m}ultilingual, \textbf{m}ulti-domain, and \textbf{m}ulti-generator corpus of \textbf{MGT}s --- M4GT-Bench. The benchmark is compiled of three tasks: (1) mono-lingual and multi-lingual binary MGT detection; (2) multi-way detection where one need to identify, which particular model generated the text; and (3) mixed human-machine text detection, where a word boundary delimiting MGT from human-written content should be determined. On the developed benchmark, we have tested several MGT detection baselines and also conducted an evaluation of human performance. We see that obtaining good performance in MGT detection usually requires an access to the training data from the same domain and generators. The benchmark is available at \url{https://github.com/mbzuai-nlp/M4GT-Bench}.
%Promising results always occur when training and test data distribute within the same domain or generators.

\end{list} % Added to suppress compile errors.
\end{abstract}

\input{section/1_intro}

\input{section/2_relatedwork}
\input{section/3_data}

\input{section/4_exp}
\input{section/5_conclusion}

\bibliography{custom}
% \bibliographystyle{acl_natbib}

% \section{Example Appendix}
% \label{sec:appendix}
\input{section/6_appendix}
% This is a section in the appendix.

\end{document}

%% file: section/1_intro.tex
\section{Introduction}
% \todo{Make multilinguality more prominent: in the abstract, intro, conclusion, discussion, etc.}
% why we do MGT detection
The advent of Large Language Models (LLMs) such as \chatgpt\footnote{We refer to GPT-3.5-Turbo throughout this paper.} and \gptfour marks a transformative era in text generation. These models are able to generate a coherent text that is very similar to human-written content. They are also easily accessible and becoming more and more widespread. While such tools significantly boost productivity across various fields such as journalism, social media, education, and academic writing, they also introduce unprecedented avenues for misuse and, consequently, pose negative societal implications. 

LLMs can generate immense amounts of deceptive fake news cluttering the information space. In social media, they have the potential to automatically generate fake accounts and increase their influence in social media communities without much effort from their owners. In the academic context, these technologies may advance beyond mere replication of existing content, potentially diminishing the effectiveness of established plagiarism detection methodologies and, consequently, posing a threat to the fundamental tenets of scholarly integrity. As a result, it could impact academic writing by leading to an increase in publications of automatically generated papers lacking scientific merit in respected journals and conferences. Finally, the scientific community has started reporting on cases of potentially automatically generated reviews in some well-known conferences.

Studies have shown that humans perform only marginally better than random chance in distinguishing human-written from machine-generated texts (MGTs)~\cite{wang2023m4}. Therefore, automatic MGT detection becomes essential in tackling misinformation, preserving the integrity of digital platforms and the scientific community, and ensuring trust in communications.

%underscoring the pressing need for the development of robust, automatic systems capable of identifying machine-generated text reliably.

%They're reshaping various domains, from journalism and social media to educational resources and academic writing. However, this rise also brings critical concerns about misuse and societal impact. 

% Distinguishing machine-generated texts from human-produced content has become vital to counter misinformation, uphold the authenticity of digital platforms and the scientific community, and maintain trust in communications.

% The advent of Large Language Models (LLMs) has brought about a shift in the landscape of textual content generation. LLMs like ChatGPT and GPT-4 have become pervasive and easily accessible, their remarkable fluency and versatility in text generation catalyze an unprecedented surge in machine-generated content across diverse communication channels, from shaping news articles to orchestrating conversations on social media, addressing queries on forums, facilitating education, and even contributing to academic discourse. It raises legitimate concerns about its potential misuse and societal implications. The need to identify and differentiate such content from genuine human-generated text is critical in combating disinformation, preserving the integrity of online platforms and scientific field, and maintaining trust in the communication.

% how previous work did in this direction
% what's the difference of our tasks and datasets between theirs
% Can Jonibek or Jinyan help?
Previous work typically frames MGT detection as binary classification \cite{zellers2019defending, mitchell2023detectgpt, bao2023fast}, focusing primarily on English. Moreover, the majority of studies also overlook the fact that the content can be a mixture of an MGT and human-written text. These limitations make the previous experimental setups significantly different from practice.

In this work, we address these issues by presenting a new benchmark for MGT detection that consists of three tasks, which were not previously explored in this exact formulation, each shedding light on different facets of this challenge.
% \todo{Change the task A, B, C to Task 1, 2, 3}
\begin{itemize}
    \item \textbf{Task 1: Binary Human-Written vs. Machine-Generated Text Classification.} The objective is to categorize a given text as either human-written or machine-generated. This task is similar to the problem formulations in the previous work. However, our dataset offers annotated resources of new domains and generators in multiple languages. Task 1 sets two distinct tracks, one focusing exclusively on English sources (monolingual), and the other embracing a multilingual scope and introducing greater diversity compared to the previous work. 
    
    \item \textbf{Task 2: Multi-Way Machine-Generated Text's Generator Detection.} This task involves identifying the specific generator responsible for producing a given text. The text may be either human-written or crafted by a LLM. A different perspective to look at Task 2 is that we are attributing authorship to a specific generator~\cite{munir2021through}. %, uchendu2020authorship, schuster2020limitations

    \item \textbf{Task 3: Human-Written to Machine-Generated Text Change Point Detection.} In this task, the goal is to precisely recognize the boundary within the mixed text, where the transition occurs from human-written to machine-generated content.
\end{itemize}

Our contributions are summarized below:
% \todo{Mention multilinguality}
% Artem, Yuxia
\begin{itemize}
    \item We construct a diverse public multilingual MGT detection benchmark involving nine languages, six domains, nine LLM generators and three different tasks. 
    \item We introduce a novel task formulation for MGT detection, where human-machine mixed text is explored. 
    It is the first attempt to evaluate the ability of automatic approaches on detecting a boundary between human-written and LLM-generated texts.
    \item On the developed benchmark, we have tested several strong MGT detection baselines and also conducted an evaluation of human performance. 
    %Low H%human performance on the Task 2 , demonstrating the challenges to distinguish unique LLMs.
    We see that obtaining good performance in MGT detection usually requires an access to the training data from the same domain and generators.
    % We analyze the human performance utilizing a few-shot learning approach\footnote{The human evaluation task was done by experts in NLP. Different performance might be obtained by average human.}.  yields a performance level below random chance in distinguishing different LLM generations, suggesting the challenges of task 2. We further demonstrate strong baselines for each task and investigate their performance in various experimental settings.
    % \item New public datasets for these tasks. (some are brand new, like for task C, some are repurposing and extensions and extensions of previous tasks.)
    % \item Strong baselines, detailed experiments and analysis.
\end{itemize}

%% file: section/2_relatedwork.tex
\section{Related Work}
% \todo{cite also https://arxiv.org/abs/2303.14822 -- they are whitebox and English}

%, where the goal is to distinguish texts generated by certain language models from those authored by humans. 
% However, with the rapid advancement of a wide range of LLMs such as GPT-4, Bard, Cohere, and Claude, they can be easily used to generate texts through simple API calls.
% However, there is a growing interest for more fine-grained classification that can not only identify the nature of the texts (i.e., \textit{whether it is machine-generated or human written}), but also its specific source (i.e., \textit{which LLM generates it?}). This is also known as authorship attribution \cite{uchendu2020authorship, venkatraman2023gpt, rivera2024few}. 
% % \cite{achiam2023gpt}

% \subsection{Binary Detection}
\paragraph{Binary Detection}
The task of detecting MGTs has traditionally been formulated as a binary classification problem \cite{zellers2019defending, gehrmann-etal-2019-gltr, solaiman2019release, ippolito2019automatic}.
MGT detection can be broadly categorized into two main types: supervised and unsupervised methods. Supervised approaches \cite{wang2023m4, uchendu2021turingbench, zellers2019defending, zhong2020neural, liu2022coco} rely on annotated datasets to train classifiers. In contrast, unsupervised methods leverage white-box features such as likelihood and log-rank \cite{solaiman2019release, ippolito2019automatic, mitchell2023detectgpt, su2023detectllm, he2023mgtbench, hans2024spotting}, or employ watermarking techniques \cite{kirch2023watermark, zhao2023protecting, zhao2023provable} to identify machine-generated text. Here, we focus on supervised approaches. 

% As a benchmark, 
\citet{wang2023m4} evaluate several supervised detectors, including RoBERTa~\cite{liu2019roberta}, XLM-R \cite{conneau2019unsupervised}, logistic regression classifier with GLTR features \cite{gehrmann2019gltr}, stylistic features~\cite{li2014authorship}, and NELA~\cite{horne2019robust} features. Similar investigation of supervised methods is also done by \citet{guo2023close, hu2023radar,  xiong2024fine}. %However, most of them only consider binary detection. 

\begin{table*}[t!]
    \centering
    \resizebox{\textwidth}{!}{
            \setlength{\tabcolsep}{3pt}
    \begin{tabular}{lccc|ccccc|c|c}
    \toprule
        \textbf{Source} & \multicolumn{3}{c|}{\textbf{Human}} & \multicolumn{5}{c|}{\textbf{Parallel Data}} & \textbf{Total} & \textbf{New test}\\
        \textbf{Domain} & \textbf{Total=} &\textbf{Upsample+} & \textbf{Parallel} & \textbf{\davinci} & \textbf{\chatgpt} & \textbf{Cohere} & \textbf{Dolly-v2} & \textbf{BLOOMz} & \textbf{Machine} & \textbf{GPT-4}\\
    \midrule
      \textbf{\outfox} & 16,272 & 13,272 & 3,000 & 3,000 & 3,000 & 3,000 & 3,000 & 3,000 & 15,000 & 3,000 \\
    \midrule
      Wikipedia & 14,333 & 11,997 & 2,336 & 3,000 & 2,995 & 2,336 & 2,702 & 2,999 & 14,032 & 3,000\\
      Wikihow   & 15,999 & 13,000 & 2,999 & 3,000 & 5,557 & 3,000 & 3,000 & 3,000 & 17,557 & 3,000\\
      Reddit ELI5 & 16,000 & 13,000 & 3,000 & 3,000 & 3,000 & 3,000 & 3,000 & 2,999 & 14,999 & 3,000\\
      arXiv abstract & 15,998 & 13,000 & 2,998 & 3,000 & 3,000 & 3,000 & 3,000 & 3,000 & 15,000 & 3,000\\
      PeerRead &  2,847 & 0      & 2,847 & 2,340 & 2,340 & 2,342 & 2,344 & 2,334 & 11,700 & 2,334\\
      Total    & 65,177 & 50,997 & 14,180 & 14,340 & 16,892 & 13,678 & 14,046 & 14,332 & 73,288 & 14,344 \\
    \bottomrule
    \end{tabular}
    }
    \caption{\textbf{Tasks 1 and 2 data statistics:} all data used for Task 1; data without upsampled human for Task 2. The first row (\outfox) and the last column (\gptfour) represent newly generated data added to the M4~\cite{wang2023m4}.}
    \label{tab:dataset}
\end{table*}

% \subsection{Multi-Class Detection} % Fine-grained
\paragraph{Multi-Class Detection}
There is a growing interest for more fine-grained classification that can not only identify the nature of the texts (i.e., \textit{whether it is machine-generated or human written}), but also its specific source (i.e., \textit{which LLM generates it?}). 
%This is also known as authorship attribution \cite{uchendu2020authorship, venkatraman2023gpt, rivera2024few}.
The fine-grained multi-class classification problem is closely related to the authorship attribution \cite{uchendu2020authorship, venkatraman2023gpt, rivera2024few}. 
\citet{munir2021through} find that texts generated by LLMs contain distinguishable signals that can be used to attribute the source of texts. \citet{uchendu2020authorship} investigate three authorship attribution problems: (1) whether two texts are produced by the same generator, (2) whether a text is generated by a machine or a human, and (3) which LLM generated the text. \citet{venkatraman2023gpt} examine if the principle that humans prefer to spread information evenly can help capturing unique signatures of LLMs and human authors. Similarly, \citet{rivera2024few} leverage representations of writing styles.

% \subsection{Authorship Obfuscation}
\paragraph{Authorship Obfuscation}
Work on authorship obfuscation 
extends beyond the binary/multi-class classification perspective by addressing the adversarial context of co-authorship between human and machine~\cite{macko2024authorship}.
Several works show that MGT detection methods are susceptible to authorship obfuscation attacks such as paraphrasing, back-translation, and human-machine collaboration mixture~\cite{crothers2022adversarial, krishna2023paraphrasing, shi2023red, koike2023outfox}. 
\citet{gao2024llm} introduces a dataset with a mixture of machine and human written texts using operations such as polishing, completing \cite{xie2023next}, rewriting~\cite{shu2023rewritelm}, adding natural noise \cite{wang2021adversarial}, and adapting~\cite{gero2022sparks}. 
\citet{kumarage2023stylometric} use stylometric signals to quantify changes in tweets and detect if and when AI starts to generate tweets.
\citet{dugan-etal-2020-roft, dugan2023realorfake} investigate human ability to detect the boundary, where machine generated continuations are from LLMs. However, they use currently outdated LLMs such as GPT2-XL. In this work, 
%we suggest a task of automatic detecting and localizing the change point from human-written segments to machine-generated segments. 
we suggest the task of automatic detection and localization of change points from human-written 
segments 
to machine-generated segments. 
%
%Compared to some previous works, our task is more challenging as we generated continuations using state-of-the-art LLMs. 
Our task is more challenging than previous works as we used state-of-the-art LLMs for generating continuations.

%% file: section/3_data.tex
\section{Datasets and Metrics}
% Using the M4 dataset as foundation, we up-sampled human-written texts for each domain to make data for subtasks A and B more balanced.
This section describes the datasets for each task and the details of their creation. The data statistics is shown in \tabref{tab:dataset}, \ref{tab:Task1-multilingual} and \ref{tab:dataset-C}.

\begin{table*}[t!]
    \centering
    \small
    \tabcolsep2pt
%    \resizebox{\textwidth}{!}{
            % \setlength{\tabcolsep}{3pt}
    \begin{tabular}{lllr|cccccc}
    \toprule
        \textbf{Source/} & \textbf{Data} & \textbf{Language} & \textbf{Total} & \multicolumn{6}{c}{\textbf{Parallel Data}}  \\
        \textbf{Domain} & \textbf{License} && \textbf{Human} & \textbf{Human} & \textbf{\davinci} & \textbf{\chatgpt} & \textbf{Jais} & \textbf{\llamatwo} & \textbf{Total}  \\
    \midrule
        Arabic-Wikipedia & CC BY-SA-3.0 & Arabic & 1,209,042 &  3,000 & -- &  3,000 & -- & & 6,000\\
    % \midrule
        True \& Fake News & MIT License & Bulgarian &   94,000 &  3,000 & 3,000 & 3,000 & -- & & 9,000 \\
    % \midrule
        Baike/Web QA & MIT license & Chinese &  113,313 & 3,000 & 3,000 & 3,000 & -- & & 9,000 \\
    % \midrule
        id\_newspapers\_2018 & CC BY-NC-SA-4.0 & Indonesian &   499,164 &  3,000 & -- &  3,000 & -- & & 6,000 \\
    % \midrule
        RuATD & Apache 2.0 license & Russian & 75,291 & 3,000 & 3,000 & 3,000 & -- & & 9,000 \\
    % \midrule
        Urdu-news & CC BY 4.0 & Urdu & 107,881 & 3,000 & -- & 3,000  & -- & & 6,000 \\
    \midrule
        \textbf{News} & Apache 2.0 & Arabic & 1,000 & 1,000 & -- & 1,000 & 100 & -- & 2,100 \\
    % \midrule
        \textbf{CHANGE-it News} & CC BY-NC-SA 4.0 & Italian & 127,402 & 3,000 & -- & -- & -- & 3,000 & 6,000 \\
        \textbf{News} & CC BY-NC-SA-4.0 & German & 10,000 & 3,000 & -- &  3,000 & -- & -- & 6,000\\
        \textbf{Wikipedia} & CC BY-SA-3.0 & German & 2,882,103 & 3,000 & -- &  3,000 & -- & -- & 6,000\\
    % \midrule
    \midrule
        \bf Total & -- & -- & 5,119,196 & 28,000 & 9,000 & 25,000 & 100 & 3,000 & 65,100\\
    \bottomrule
    \end{tabular}
%    }
    \caption{\textbf{Task 1 Multilingual} introduced new languages: German, Italian, news for Arabic by \chatgpt and Jais-30B. \llamatwo-70B used here for generating Italian texts is a fine-tuned Italian version, named \textit{camoscio-70B}.}
    \label{tab:Task1-multilingual}
\end{table*}
% \todo[inline]{Add a table for multilingual Task 1}
% \subsection{Task 1: Human vs. Machine}
\subsection{Human vs. Machine}
The task 1 aims to distinguish human-written from machine-generated text --- a binary classification task.
Based on the M4 dataset~\citep{wang2023m4}, given a domain, we up-sampled human text to match the total number of machine-generated text to avoid data imbalance.
After upsampling, the dataset consists of 65,177 human-written texts and 73,288 machine-generated texts as shown in \tabref{tab:dataset}.
We additionally generated texts using GPT-4 for each domain to evaluate the detectors' generalization ability on unseen and strong generators.

For the multilingual track, we integrate new languages (German and Italian), and news for Arabic by generators of \chatgpt and Jais-30B, as the statistics shown in \tabref{tab:Task1-multilingual}.

%We perform the same data cleaning steps as in M4:
\paragraph{Data Cleaning} Simple artifacts in human-written texts, such as multiple newlines and bullet points, could assist detectors. Their presence in the training data may discourage detectors from learning more generalized signals.
Most importantly, these artifacts primarily originate from data crawling (multiple newlines at the beginning and the end of text) and conversion (such as PDF to text).

To recover the original format of human-written plain text,
%--- the text is closer to what a human directly writes,
% Therefore, 
we performed minimal cleaning on the upsampled human-written texts: 
(\emph{i})~in a \wikihow text, we removed multiple commas at the beginning of a new line (e.g. ``,,,,,,, we believe that ...'') and repeating newlines (``$\backslash$n$\backslash$n$\backslash$n$\backslash$n \textit{text begin} $\backslash$n$\backslash$n$\backslash$n$\backslash$n”);
% (\emph{ii})~in machine-generated \wikihow texts, we removed bullet points (as there were no bullet points in human-written texts);
(\emph{ii})~in \wikipedia articles, we removed references (e.g., [1], [2]), URLs, multiple newlines, as well as paragraphs whose length was less than 50 characters; and 
(\emph{iii})~in \arxiv abstracts, we removed newlines stemming from PDF conversion.

\paragraph{Metrics} In addition to accuracy, we also use precision, recall, and F1-score with respect to MGT. % to evaluate detectors.
% Answer the following questions:
% (1) based on M4, how did we get subtask A training and dev sets? 
% (2) what's the statistic distribution for training and dev sets?
% (3) what tests do we plan to have?
% (4) the experiment settings, and why we highlights unseen generators in subtask A?
% (5) in multilingual setting, how did we do? what kind of experiments we can do?

% \subsection{Task 2: Multi-way Detection}
\subsection{Multi-way Detection} 
Task 2 is to determine which generator model produces the considered text. 
This is motivated by the Intelligence Protection (IP) of LLMs generations. For example, it can be used to detect whether \emph{model A} generations are extensively used to train another commercial \emph{model B} by distillation (this is forbidden by OpenAI). If many generations from \emph{B} are detected as \emph{model A} generations, one might suspect that \emph{model B} used a large amount of generations of \emph{model A} in the training process.

In general, identifying the specific model used to produce a copyrighted material is important for legal and copyright reasoning.  Additionally, in cases where the generated material is harmful, misleading, or illegal, pinpointing the exact LLM responsible is essential for addressing ethical concerns and legal obligations.

Different from task 1, we use the parallel subset for task 2 without the upsampling of human-written texts, including six generators: \chatgpt, \davinci, \gptfour, \cohere, \dolly, and \bloomz.
A new domain \outfox is collected to evaluate the classifier's domain generalization in student essays. % based on a problem statement .

\paragraph{Metrics} 
F1-score, precision, recall in terms of seven individual labels representing different generators and human-written texts are used. We also provide F1-macro and accuracy.

\begin{table*}[t!]
    \centering
    \small
    %\tabcolsep2pt
   % \resizebox{\columnwidth}{!}{
    \begin{tabular}{ll | llll}
    \toprule
    \textbf{Domain} & \textbf{Generator} & \textbf{Train} & \textbf{Dev} & \textbf{Test} & \textbf{Total} \\
    \midrule
    \multirow{5}{*}{\peerread}
    & ChatGPT & 3,649 (232) & 505 (23) & 1,522  (89) & 5,676 (344) \\
    & \llamatwo-7B* & 3,649 (5) & 505 (0) & 1,035 (1) & 5,189 (6) \\
    % \midrule
    & \llamatwo-7B & 3,649 (227) & 505 (24) & 1,522  (67) & 5,676 (318)\\
    & \llamatwo-13B & 3,649 (192) & 505 (24) & 1,522  (84) & 5,676 (300) \\
    & \llamatwo-70B & 3,649 (240) & 505 (21) & 1,522  (88) & 5,676 (349) \\
    \midrule
    \multirow{5}{*}{\outfox}
    % & ChatGPT & \\
    & GPT-4 & -- & -- & 1,000 (10) & 1,000 (10) \\
    & LLaMA2-7B & -- & -- & 1,000 (8) & 1,000 (8) \\
    & LLaMA2-13B & -- & -- & 1,000 (5) & 1,000 (5) \\
    & LLaMA2-70B & -- & -- & 1,000 (19) & 1,000 (19) \\
    \bottomrule
    \end{tabular}
    %}
    \caption{\textbf{Task 3 boundary identification data} based on GPT and \llamatwo series over domains of academic paper review (\peerread) and student essay (\outfox). The number in ``()'' is the number of examples purely generated by LLMs, i.e., human and machine boundary index=0. \llamatwo-7B* and \llamatwo-7B used different prompts.}
    \label{tab:dataset-C}
\end{table*}

% \subsection{Task 3: Boundary Identification}
\subsection{Boundary Identification}
% why we choose peerread for subtask C
Task 3 aims to imitate the real-world LLM usage situation, where not the full text is generated by machine, but partially written by human and partially generated by a machine. To simplify the task, we formulate it as first written by a human and then continued by a machine, and the task is to detect the single boundary of change. 

We consider two common misuse scenarios of human-machine mixed text --- continuing to write academic paper reviews and student essays.
% how did we generate data using chatgpt? what's the challenges in data generation? % what prompts do we use?
Given the partial human-written essay with the corresponding problem statement and the partial human-written paper reviews with the title and abstract, we complete them by generating text based on GPT and \llamatwo series using prompts as shown in \figref{fig:subtaskC-prompt}.
The proportion of words that are human-written range from 0 to 50\%.
% (4) what the input and the expected predictions, particularly the definition about the whitespace or the separating symbols which participants are confusing
The goal of a detection model is to identify the boundary position from which the text is generated by machines.

Mixed texts are specifically generated using \chatgpt, \gptfour, and \llamatwo (7B, 13B and 70B). We generate 5,676 and 1,000 examples for two domains respectively using each generator (full statistics is presented in \tabref{tab:dataset-C}).

\paragraph{How realistic is the task setting?}
%We find that it is a quite common use-case to provide a beginning of the text and ask an LLM to generate the rest. 
Providing the beginning of a text and asking an LLM to generate the rest is a common use case.
%
%At the same time, we consider that MGTs seriously modified by humans might not represent a malignant use case of LLMs, since LLMs are actually legit writing tools. 
At the same time, we believe that MGTs seriously modified by humans might not represent a malicious use case of LLMs, as LLMs are legitimate writing tools.
Depending on the amount of human intervention, one could argue that this is not anymore machine-generated, but machine-human collaboration. Therefore, we seek balance between the complexity of the task and its practicality. 
Moreover, detecting texts that contain multiple changing points is an interesting research direction for future work, but it would bring new challenges with analyzing shorter spans. Extremely, if a human added an extra word in a machine-generated sentence, it would be much harder for blackbox detectors to identify this word. %, in which whitebox setups might be the solution.
% We will increase the complexity to include human/machine polishment and multiple changing points in the future work.

% (5) metrics used here, and why we use these two metrics? according to what previous work or other reasons?
\paragraph{Metrics}
Mean Absolute Error (MAE) is used to evaluate the boundary detection model's performance. It measures the average absolute difference between the predicted position index and the actual changing point.

\begin{figure}[t!]
	\centering
 % https://docs.google.com/spreadsheets/d/1wLPDhpjA3q1D1xTH0cwQ0mdsQ-FmJjwcw8vEqriYrMA/edit?usp=sharing
 	\includegraphics[width=\columnwidth]{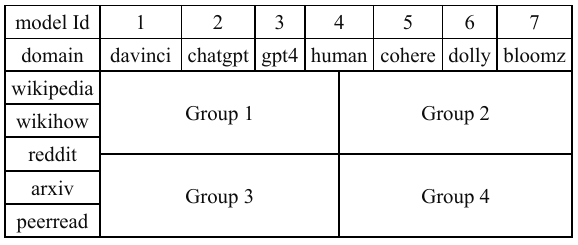}
 \caption{We split 140 examples into four groups, each involving three domains and four generators, with 48 examples including five demonstrations for learning. }
 	\label{fig:human-eval-data-sample}
\end{figure}
\section{Human Evaluation}
We investigate the complexity of differentiating the language model used for creating machine-generated texts for human readers. 
Considering that annotators may lose patience to learn and summarize the patterns of seven distinct generators, we control the number of unique models they could identify and simplify to four-class classification with five demonstration examples (five-shot) as practice to learn before the formal annotation.

% \subsection{Data for Human Evaluation}
\paragraph{Data for Human Evaluation}
There are 35 unique domain-generator combinations --- five domains $\times$ seven generators (six LLMs plus human-written texts) for task 2 data.
With the concern of annotators’ workload, we sample 4 examples for each combination, in total of 140 examples, split into four groups as \figref{fig:human-eval-data-sample}.

% Considering domain, 
Without consideration of out-of-domain classification, we follow the best-performing setup: see all, classify all.
If each annotator needs to check all domains, they need to learn 20 demonstration examples, which is a time consuming and laborious undertaking. Let us consider that each annotator annotates 3 domains. They would need to check 4 models $\times$ 3 domains $\times$ 4 examples = 48, which is in a more reasonable range.
% (1 set for learning and 3 sets for formal classification)

Five demonstration examples are randomly selected, to ensure that four classes are all included.\\
\textbf{Annotator background.}
Group 1, 2, 3, and 4 are annotated by native Italian, Chinese, English, and Russian speakers, respectively; they are either postdoc or PhD students in NLP.
% Group1: Yellow by Giovanni
% Group2: Blue by Jinyan
% Group3: Red by Tarek
% Group4: Green by Jonibek.
\begin{table}[t!]
    \centering
    \small 
    \tabcolsep5pt 
 %   \resizebox{0.95\columnwidth}{!}{
    \begin{tabular}{c|cccc}
        \toprule
      % \multirow{2}{*}{Annotator}&  \multicolumn{4}{c}{Overall Performance} & \multicolumn{7}{c}{Separate class F1-score} \\
       \textbf{Annotator} & \textbf{Prec} & \textbf{Recall} & \textbf{F1-macro} & \textbf{Accuracy} \\ % & Human & \davinci & \chatgpt & \gptfour & \cohere & \dolly & \bloomz \\
       \midrule
         Group 1&      \textbf{27.91} & \textbf{28.25} & \textbf{28.18} &\textbf{ 27.42} \\ % & 12.50 & 32.00 & 23.08 & 42.11 & --- & --- & --- \\
         Group 2&      16.28 & 09.22 & 12.91 & 10.45 \\ % & 32.26 & --- & --- & --- & 10.00 & 10.00 & 0.00 \\
         Group 3&      20.93 & 16.00 & 16.91 & 15.47 \\ % & 32.26 & --- & --- & --- & 9.52 & 13.33 & 22.22 \\
         Group 4&      25.58 & 25.83 & 25.25 & 24.82 \\ % & 25.00 & 47.62 & 0.00 & 26.67 & --- & --- & --- \\
         \midrule
         Group 1 + 2 & 22.09 & 19.73 & 21.10 & 20.27 \\ % & 25.53 & 32.00 & 22.22 & 42.11 & 10.00 & 10.00 & 0.00  \\
         Group 3 + 4 & 23.26 & 22.98 & 20.84 & 21.13 \\ % & 28.57 & 47.62 & 0.00 & 26.67 & 9.52 & 13.33 & 22.22 \\
         All &         22.67 & 22.11 & 21.06 & 21.20 \\ % & 27.27 & 39.13 & 13.04 & 35.29 & 9.76 & 11.43 & 12.50\\
    \bottomrule
    \end{tabular}
%    }
   \caption{\textbf{Task 2 human evaluation performance.} Each group performs a four-class classification task. Results are even worse than random guess (25\%).}
    \label{human-eval}
\end{table}

% \subsection{Results}
\paragraph{Results}
Intuitively, it would be challenging to distinguish between generators based on 1-shot per class.
Annotators may make decisions depending on the observations that generations by \cohere tend to be repetitive, \dolly has high-quality and coherent texts, while \bloomz texts are more likely to contain a bunch of numbers or a bunch of ``YES'' or ``NO'' characters repetitively at the end of the texts, compared to other models. 
% confusing and require spending more time reading to understand.
For GPT series, annotators distinguish them depending on their quality level assumption that \gptfour > \chatgpt > \davinci.
They spot human-written texts based on formatting patterns (e.g., initial double newlines, initial space, completely missing new lines in the paragraph), typos, inconsistencies within texts, or specific references and URLs. 

As shown in \tabref{human-eval}, the best distinction accuracy for humans is 27\%, and the average accuracy 21\% is less than random guess.
This implies that it is extremely difficult for humans to learn patterns from five demonstration examples and then distinguish generations of different LLM generators (see the F1-score for each separate class in \appref{app:humaneval}).

%% file: section/4_exp.tex
\section{Experiments}
% For three tasks, we provide results with several strong baselines.
We conduct an experimental evaluation with several common baselines representing a diverse set of detection techniques: standard fine-tuning of an encoder-based Transformer model, elaborated feature engineering for the similar task, ranking-based features which form patterns for different LLMs. For classification tasks 1 and 2, we use five baseline classifiers suggested by \citet{wang2023m4}. They include fine-tuned RoBERTa and XLM-R classifiers, logistic regression with GLTR features (LR-GLTR)~\cite{gehrmann-etal-2019-gltr}, SVM with Stylistic (Stylistic-SVM), and SVM with NELA features (NELA-SVM)~\citep{horne2019robust}. For the boundary detection task 3, we applied two detectors based on semantic features.
%that are common for the tasks of such type, without applying exceptionally-novel approaches. 
%This mainly aims to investigate the lower bound of the detection performance using simple but diverse techniques. 

%We applied two detectors for boundary identification based on semantic features.
%For classification tasks 1 and 2, we use five baseline classifiers suggested by \citet{wang2023m4}. They include fine-tuned RoBERTa and XLM-R classifiers, logistic regression with GLTR features (LR-GLTR), SVM with Stylistic (Stylistic-SVM), and SVM with NELA features (NELA-SVM).

%The selected baselines implement different detection principles: standard fine-tuning of an encoder-based Transformer model, elaborated feature engineering for the similar task, ranking-based features which form patterns for different LLMs. For the boundary detection task 3, we applied two detectors based on semantic features.
%\footnote{They are selected based on different principles: standard fine-tuning of an encoder-based Transformer model, elaborated feature engineering for the similar task, ranking-based features which form patterns for different LLMs.}
%The results demonstrate the difficulty of these tasks.

% \paragraph{RoBERTa and XLM-R}
% We fine-tune two Transformer-based models for MGT detection: RoBERTa and XLM-R.
%\tabref{tab:subtaskA} presents accuracy, precision, recall, and F1 scores with respect to machine generations averaged over 5 runs. 
% TODO: anything else?
% TODO: what are hyperparameters?

\paragraph{GLTR features} 
These features
are based on the assumption that in order to generate fluent and natural-looking text, most LLM decoding strategies sample high-probability tokens from the head of the distribution~\cite{gehrmann-etal-2019-gltr}. Thus, word ranking information of an LM can be used to distinguish machine-generated texts from human-written. 
Following \citet{wang2023m4}, we select two categories of these features: (1) the number of tokens in the top-10, top-100, top-1000, and 1000+ ranks from the LM predicted probability distributions (4 features); and (2) the Frac(p) distribution over 10 bins ranging from 0.0 to 1.0 (10 features). Frac(p) describes the fraction of probability for the actual word divided by the maximum probability of any word at this position.
We train a logistic regression model based on these 14 features to perform the binary classification task with the parameter of the maximum iteration = 1,000.

\paragraph{Stylistic and NELA features}
% Following \citet{wang2023m4}, 
We use stylistic features including (1) character-based features, e.g., the number of characters, letters, special characters, etc., (2) syntactic features, e.g., a number of punctuation and function words, (3) structural features, e.g., a total number of sentences, and (4) word-based features, e.g., a total number of words, average word length, average sentence length, etc.

NEws LAndscape (NELA) features~\citep{horne2019robust} involves six aspects: (1) style: the style and the structure of the article; (2) complexity: how complex the writing is; (3) bias: overall bias and subjectivity; (4) affect: sentiment and emotional patterns; (5) moral based on the Moral Foundation Theory~\citep{grahm2012hadit}, and (6) event: time and location. An SVM classifier is applied to perform the binary distinction.

\begin{table}[t!]
    \centering
    \small 
    % \scalebox{1.}{
 %   \resizebox{\columnwidth}{!}{
    % \setlength{\tabcolsep}{1.5pt}
\tabcolsep3pt 
    %\resizebox{\columnwidth}{!}{
    \begin{tabular}{l|l|cccc}
        \toprule
        \textbf{Detector} & \textbf{Test} & \textbf{Prec} & \textbf{Recall} & \textbf{F1} & \textbf{Acc} \\
        \midrule
            \multirow{7}{*}{RoBERTa} 
             & All & 99.16 & 99.56 & \textbf{99.36} & \underline{\textbf{99.26}} \\
             & \davinci & 71.08 & 98.53 & 82.58 & 79.21 \\
             & \chatgpt & 74.64 & 99.93 & 85.45 & 82.99 \\
             & \gptfour & 70.81 & 100.00 & 82.90 & 79.37 \\
             & \cohere & 70.11 & 98.50 & 81.91 & 78.24 \\
             & \dolly & 69.95 & 97.46 & 81.44 & 77.78 \\
             & \bloomz & 60.31 & 60.16 & 60.22 & 60.30 \\
            % \cline{1-6}
            \midrule
            \multirow{7}{*}{XLM-R} 
             & All & 95.08 & 98.80 & \textbf{96.87} & \textbf{96.31} \\
             & \davinci & 80.57 & 90.46 & 85.23 & \underline{84.32} \\
             & \chatgpt & 78.12 & 99.95 & 87.57 & 85.62 \\
             & \gptfour & 69.44 & 99.93 & 81.93 & 77.95 \\
             & \cohere & 79.59 & 97.98 & 87.74 & 86.23 \\
             & \dolly & 76.77 & 84.58 & 80.40 & \underline{79.43} \\
             & \bloomz & 73.98 & 72.16 & 72.74 & \underline{73.07} \\
            \midrule
            \multirow{7}{*}{GLTR-LR} 
             & All & 84.59 & 88.71 & 86.60 & 84.26 \\
             & \davinci & 81.62 & 78.13 & 79.83 & 80.27 \\
             & \chatgpt & 82.00 & 96.33 & 88.59 & \underline{87.59} \\
             & \gptfour & 83.07 & 98.17 & 89.99 & \underline{89.08} \\
             & \cohere & 82.98 & 99.27 & \textbf{90.40} & \underline{\textbf{89.46}} \\
             & \dolly & 81.10 & 72.24 & 76.42 & 77.70 \\
             & \bloomz & 76.02 & 50.45 & 60.65 & 67.27 \\
            \midrule
            \multirow{7}{*}{Stylistic-LR} 
             & All & 86.42 & 76.67 & \textbf{81.25} & \textbf{84.91} \\
             & \davinci & 81.14 & 47.16 & 59.65 & 68.10 \\
             & \chatgpt & 65.96 & 50.67 & 57.31 & 62.26 \\
             & \gptfour & 97.62 & 44.82 & 61.44 & 71.87 \\
             & \cohere & 75.67 & 44.18 & 55.79 & 64.99 \\
             & \dolly & 77.75 & 49.37 & 60.39 & 67.62 \\
             & \bloomz & 57.93 & 48.86 & 53.01 & 56.69 \\
             \midrule
            \multirow{7}{*}{NELA-LR} 
             & All & 74.55 & 63.78 & 68.75 & 75.27 \\
             & \davinci & 77.52 & 60.25 & 67.80 & 71.39 \\
             & \chatgpt & 82.12 & 59.68 & 69.12 & 73.34 \\
             & \gptfour & 93.30 & 56.60 & \textbf{70.46} & \textbf{76.27} \\
             & \cohere & 74.82 & 58.32 & 65.55 & 69.35 \\
             & \dolly & 64.78 & 62.28 & 63.50 & 64.21 \\
             & \bloomz & 44.64 & 77.58 & 56.67 & 40.69 \\      
        \bottomrule
    \end{tabular}
%}
    \caption{\textbf{Task 1 monolingual binary human vs. machine performance} on a test generator. Accuracy (Acc), Prec (precision), Recall, and F1-scores(\%) \textbf{with respect to machine-generated text}. The classifier was trained on the data of all generators except for the test generator. \textit{All} refers to the setting that randomly splits train. validation and test sets, each has data of all generators.
    }
    \label{tab:subtaskA}
\end{table}

% \subsection{Task 1: Monolingual} 
\subsection{Monolingual Human vs. Machine} 
\label{sec:subtaskA-mono}
For Task 1, we simulate a practical scenario where a detector have to deal with texts generated from a new LLM while it is trained on the outputs from a limited number of other generators. A practical detector should be able to generalize over different generators and be robust on unseen inputs.

\paragraph{Experimental Setup}
We conduct experiments in two setups.
(1) In the first setup, we select an LLM and combine its generations across all domains with the upsampled human-written texts to create a test set. Texts produced by other LLMs and humans are divided into training and development sets with a 4:1 ratio. 
(2) In the second setup, we take the data from all domains and generators and randomly split it into train and test sets (row All in \tabref{tab:subtaskA}).
% Evaluation metrics in this task are accuracy, precision, recall, and F1 score.  
% In all experiments, 
We train detectors with five different seeds for all experiments and 
%[0, 10, 30, 55, 75] 
present the mean values of the metrics.

\paragraph{Results and Analysis}
Classification results of five detectors are presented in \tabref{tab:subtaskA}.
We analyze the results to answer three research questions: (1) For a given detector, on which unseen generator does it generalize best (highlighting F1-score and accuracy)? (2) Given a setting that training without the data of the generator to test, which detector performs the best (underlined Acc). (3) On average across both settings, which detector is the most accurate and has the best generalization capabilities? Which detector has higher recall than precision? %, what does this imply?

RoBERTa, XLM-R, and LR based on stylistic features give the best F1-score and accuracy on the setting where the classifier is trained with generations of all generators and tested across all generators.
GLTR features perform the best in the setting where the detector is trained on data without generations from \cohere, and tested over Human vs. \cohere data, while NELA has the highest accuracy in distinguishing human vs. \gptfour when it has not seen any data from \gptfour. 
%\todo[]{why?}

The results for RoBERTa and XLM-R suggest that such detectors encountered notable challenges when confronted with text from new generators not encompassed in their training data, exhibiting the large accuracy gap between the setting of \textit{All} and others. 
This problem is especially conspicuous in the case of \bloomz for all detectors (All vs. \bloomz), followed by \dolly, \davinci, \chatgpt, \gptfour, and \cohere.
% Among all generators, generation of \bloomz is the most difficult one to identify without seeing during training, followed by \dolly, \davinci, \chatgpt, \gptfour and \cohere.
XLM-R and GLTR demonstrate better generalization performance (underlines numbers) over (\davinci, \dolly, and \bloomz) and (\chatgpt, \gptfour, and \cohere) respectively. 
% \todo[]{check that all mentions of \dolly, \gptfour are replaced by latex commands}

% \todo[]{Add macro averaged results in the table}
% We do not add macro because task 1 English is a setting of balanced data M vs H, accuracy is similar to macro.

\textbf{XLM-R is the best on average.} It has higher accuracy and better generalization ability. LR with NELA features is the worst.
Recall of RoBERTa, XLM-R, and GLTR is higher than precision, while it is the opposite for stylistic and NELA features. 
%Recall in terms of machine-label reflects the ratio of true predictions over the real machine text, precision measures the ratio of true predictions over all examples predicted to be machine-generated text.
In the scenario of detecting misuse, we expect higher recall to avoid less misses of potentially malice. %ious use.

Overall, detectors show poor generalization performance when testing on unseen generators. This is well-known due to exposure bias. The generations by \bloomz are significantly different from other generators, leading to the result that when training on the data combination of other generators and testing on \bloomz, the performance is extremely low for all detectors. The detector based on NELA features overall performs the worst, which demonstrates that the features used for fact verification are not suitable to serve as signals to distinguish human vs. machine text.

\begin{table}[t!]
    \centering
    \small
    \tabcolsep4pt 
  %  \resizebox{\columnwidth}{!}{
    \begin{tabular}{l|ccc|cc}
        \toprule
        \textbf{Test} & \textbf{Prec} & \textbf{Recall} & \textbf{F1-score} & \textbf{Acc} & \textbf{F1-macro}\\
        \midrule
        All & 91.49 & 98.52 & 94.86 & 94.52 & 94.49\\
        Arabic & 88.91 & 97.35 & 92.64 & 92.18 & 92.12\\
        Bulgarian & 51.41 & 99.97 & 67.90 & 52.74 & \textbf{39.18} \\
        Chinese & 76.85 & 95.18 & 84.73 & 82.51 & 81.93 \\
        English & 63.72 & 84.64 & 72.42 & 66.51 & 64.55\\
        German & 66.21 & 98.30 & 79.00 & 73.62 & 71.58\\
        Indonesian & 53.13 & 100.0 & 69.39 & 55.83 & \textbf{45.00}\\
        Italian & 75.20 & 100.0 & 85.84 & 83.51 & 83.04 \\
        Russian & 52.55 & 86.38 & 65.26 & 53.70 & \textbf{47.60}\\
        Urdu & 91.29 & 97.93 & 94.49 & 94.39 & 94.39 \\
        \bottomrule
    \end{tabular}
%}
    \caption{\textbf{Task 1 multilingual human vs. machine \xlmr} Prec (precision), Recall, and F1-scores(\%) with respect to \textbf{MGT} on a test language. The classifier was trained on the data of all languages except for the test language. \textit{All} refers to the setting that randomly splits train. validation and test sets over all languages.
    }
    \label{tab:subtaskA_multilingual}
\end{table}

\subsection{Multilingual Human vs. Machine}
\label{sec:subtaskA-multi}
Similar to monolingual setups, we evaluate detectors generalization on new languages (the language to test) and train on the combination of other languages. In the setting of \textit{All} in \tabref{tab:subtaskA_multilingual}, we merge data of all languages, and randomly split into train, validation and test sets. 

\paragraph{Results and Analysis}
We categorize the performance of the detector (\xlmr) across ten configurations into three tiers based on the F1 score.
(1) F1<50: Bulgarian, Russian and Indonesian;
(2) 60<F1<85: English, German, Italian, Chinese; and
(3) F1>90: Arabic, Urdu, All.

For the first level, Bulgarian and Russian are both Slavic languages, and Indonesian is a member of the Malayo-Polynesian branch of the language family, training on other distantly-unrelated languages (Latin, Chinese and Arabic family) would naturally result in low performance.
For the second level, Chinese is a high-resource language, so the detector could benefit from the learned patterns during \xlmr pretraining. 
For German and Italian, English training data will keep the same family languages in a medium performance, vice versa.

High recall and low precision with respect to MGT across all testing languages implies that the majority of examples are recognized as machine-generated text.
This leads to lower accuracy on human-written text than that on MGT (unweighted average macro-F1 < MGT F1).

Overall, the detector often predicts texts as MGT.
when dealing with low-resource languages or training on language families that are distantly related, the accuracy tends to be moderate.

\begin{table}[t!]
    \centering
     \small
    \tabcolsep2.5pt 
 %   \resizebox{\columnwidth}{!}{
    \begin{tabular}{c|c|cccc}
        \toprule
        \multirow{2}{*}{\textbf{Detector}} & \textbf{Test} & \multicolumn{4}{c}{\textbf{Overall Performance}} \\ % & \multicolumn{7}{c}{Separate class F1-score} \\
        & \textbf{Domain} & \textbf{Prec} & \textbf{Recall} & \textbf{F1-macro} & \textbf{Acc}   \\%& Human & \davinci & \chatgpt & \gptfour & \cohere & \dolly & \bloomz \\
        \midrule
        \multirow{7}{*}{RoBERTa}
        & All & 96.96 & 97.01 & \textbf{96.94} & \textbf{97.00} \\ % & 96.34 & 94.43 & 96.38 & 96.74 & 97.12 & 93.30 & 99.31 \\
        & \arxiv & 55.72 & 36.55 & \underline{32.29} & \underline{36.55} \\ %& 44.91 & \textcolor{red}{0.49} & 27.97 & 64.61 & 2.28 & 38.91 & 56.49 \\
        & \peerread & 70.58 & 70.12 & 66.89 & 69.47 \\ % & 58.71 & \textcolor{red}{0.47} & 66.45 & 53.79 & 62.60 & 94.25 & 94.60 \\
        & \reddit & 77.21 & 74.49 & 71.66 & 74.49 \\ % & 89.89 & 10.38 & 50.14 & 29.12 & 84.72 & 77.12 & 94.92 \\
        & \wikihow & 72.62 & 70.56 & 68.85 & 68.36 \\ % & 27.95 & 39.25 & 50.45 & 86.94 & 67.46 & 83.72 & 98.57 \\
        & \wikipedia & 46.37 & 52.22 & 39.37 & 51.91 \\ % & 0.22 & \textcolor{red}{1.02} & 0.53 & 64.30 & 64.67 & 55.78 & 76.74 \\
        & \outfox & 71.71 & 65.04 & 66.40 & 78.25\\
        \midrule
        \multirow{7}{*}{XLM-R}
        & All & 90.73 & 90.37 & \textbf{90.16} & \textbf{90.17} \\ % & 91.41 & 80.63 & 87.48 & 90.30 & 81.60 & 72.03 & 98.75 \\
        & \arxiv & 51.29 & 43.88 & 41.71 & 43.88  \\ %& 52.99 & 10.15 & 73.11 & 7.06 & 39.07 & 40.30 & 49.08 \\
        & \peerread & 53.68 & 52.14 & 46.10 & 50.91 \\ % & 17.19 & 0.53 & 66.21 & 32.68 & 38.51 & 62.64 & 96.13 \\
        & \reddit & 69.73 & 58.72 & 57.34 & 58.71 \\ % & 75.21 & 48.82 & 49.27 & 37.89 & 58.66 & 52.55 & 97.02 \\
        & \wikihow & 65.73 & 60.84 & 58.45 & 57.38 \\ % & 43.67 & 55.08 & 26.79 & 64.95 & 66.35 & 46.13 & 98.67 \\
        & \wikipedia & 60.04 & 42.55 & \underline{38.80} & \underline{41.95} \\ % & 8.42 & 40.51 & 31.51 & 36.79 & 38.28 & 48.21 & 61.70 \\
        & \outfox &  51.94 & 42.44 & 43.00 & 52.10\\

        % without outfox
        % \midrule
        % \multirow{6}{*}{GLTR-LR}
        % & All & 42.9 & 46.11 & 43.07 & 45.84 \\ % & 66.98 & 13.15 & 37.06 & 30.17 & 50.30 & 33.83 & 70.00 \\
        % & \arxiv & 21.5 & 30.61 & \underline{23.19} & \underline{30.60} \\ % & 55.28 & 0.31 & 1.45 & 3.24 & 30.07 & 32.22 & 39.80 \\
        % & \peerread & 43.50 & 46.24 & 42.52 & 46.37 \\ % & 58.77 & 1.05 & 36.18 & 50.92 & 43.31 & 39.51 & 67.90 \\
        % & \reddit & 46.88 & 47.91 & \textbf{43.90} & \textbf{47.91} \\ % & 84.05 & 22.77 & 19.58 & 35.48 & 62.80 & 13.20 & 69.43 \\
        % & \wikihow & 43.87 & 38.78 & 38.28 & 37.23 \\ % & 57.49 & 14.69 & 28.41 & 11.35 & 54.79 & 18.79 & 82.44 \\
        % & \wikipedia & 41.91 & 35.66 & 35.54 & 34.08 \\ % & 67.54 & 11.78 & 14.10 & 20.12 & 32.22 & 33.90 & 69.14 \\
        
        \midrule
        \multirow{6}{*}{GLTR-LR}
        & All & 42.36 & 43.96 & 40.32 & 45.06 \\ % & 65.13 & 11.98 & 40.22 & 14.39 & 44.88 & 33.0 & 72.65 \\
        & \arxiv & 26.24 & 34.45 & \underline{26.92} & \underline{34.45} \\ % & 58.38 & 0.06 & 17.78 & 0.34 & 37.59 & 35.84 & 38.47 \\
        & \peerread & 42.20 & 44.10 & 39.04 & 44.32 \\ % & 60.13 & 0.49 & 48.37 & 19.08 & 37.83 & 40.61 & 66.77 \\
        & \reddit & 45.54 & 46.28 & \textbf{41.7} & \textbf{46.28} \\ % & 85.06 & 24.08 & 22.44 & 19.77 & 55.74 & 14.6 & 70.23 \\
        & \wikihow & 41.86 & 39.39 & 38.24 & 38.74 \\ % & 59.14 & 13.92 & 32.06 & 1.03 & 57.84 & 19.52 & 84.16 \\
        & \wikipedia & 41.62 & 36.95 & 34.62 & 35.18 \\ % & 67.62 & 8.15 & 16.62 & 0.7 & 33.41 & 35.59 & 80.25 \\
        & \outfox & 29.68 & 32.38 & 29.18 & 38.78 \\ % & 49.79 & 12.24 & 37.5 & 0.0 & 1.71 & 17.51 & 85.49 \\

        % without outfox
        % \midrule
        % \multirow{6}{*}{GLTR-SVM}
        % & All & 53.19 & 53.47 & \textbf{52.50} & \textbf{53.75} \\ % & 70.15 & 32.16 & 53.37 & 42.57 & 53.57 & 41.41 & 74.28 \\
        % & \arxiv & 27.88 & 35.40 & \underline{29.25} & \underline{35.40} \\ % & 61.14 & 0.06 & 42.95 & 0.00 & 28.55 & 33.7 & 38.33 \\
        % & \peerread & 33.98 & 37.18 & 34.27 & 37.54 \\ % & 60.61 & 0.11 & 31.57 & 1.29 & 31.45 & 41.64 & 73.21 \\
        % & \reddit & 46.68 & 47.84 & 44.73 & 47.84 \\ % & 83.34 & 14.95 & 43.89 & 18.88 & 56.88 & 15.58 & 79.62 \\
        % & \wikihow & 43.88 & 42.36 & 40.52 & 37.84 \\ %& 66.13 & 22.54 & 1.15 & 23.16 & 64.09 & 23.65 & 82.93 \\
        % & \wikipedia & 43.62 & 37.96 & 35.69 & 37.18\\ % & 53.55 & 17.18 & 40.28 & 43.48 & 32.03 & 36.89 & 26.45 \\
        \midrule
        \multirow{6}{*}{GLTR-SVM}
        & All & 52.81 & 48.42 & \textbf{47.24} & \textbf{50.39} \\ % & 69.35 & 21.39 & 49.47 & 31.21 & 46.84 & 36.04 & 76.35 \\
        & \arxiv & 22.57 & 32.29 & \underline{25.73} & 32.28 \\ %& 61.34 & 0.0 & 26.36 & 0.0 & 27.79 & 28.63 & 35.99 \\
        & \peerread & 34.10 & 39.10 & 34.81 & 39.40 \\ %& 60.64 & 0.15 & 41.71 & 0.17 & 30.74 & 36.98 & 73.26 \\
        & \reddit & 43.21 & 46.19 & 40.94 & 46.19 \\ %& 84.84 & 0.6 & 38.21 & 9.83 & 53.87 & 19.74 & 79.53 \\
        & \wikihow & 44.13 & 39.01 & 38.27 & 36.18 \\ %& 64.59 & 19.73 & 13.6 & 5.61 & 58.81 & 24.26 & 81.26 \\
        & \wikipedia & 34.86 & 30.95 & 26.72 & 29.81 \\ %& 51.83 & 2.62 & 34.37 & 0.0 & 30.19 & 37.92 & 30.09 \\
        & \outfox & 26.93 & 28.29 & 26.45 & \underline{28.58} \\ %& 51.25 & 27.8 & 2.84 & 0.0 & 0.54 & 15.71 & 86.99 \\

        \midrule
        \multirow{6}{*}{Stylistic-SVM}
        & All & 78.95 & 37.10 & \textbf{47.31} & \textbf{35.26} \\ %& 45.16 & 18.85 & 46.54 & 56.60 & 48.02 & 26.48 & 93.36 \\
        & \arxiv & 44.89 & 8.52 & \underline{12.71} & \underline{8.27} \\ %& 6.39 & 0.00 & 2.03 & 2.99 & 3.45 & 9.87 & 56.10 \\
        & \peerread & 50.72 & 21.96 & 25.43 & 20.44 \\ % &17.27 & 0.58 & 6.15 & 60.72 & 5.64 & 33.37 & 89.01 \\
        & \reddit & 60.98 & 27.25 & 31.16 & 24.43 \\ %& 50.18 & 15.22 & 10.78 & 26.11 & 29.06 & 12.92 & 87.18 \\
        & \wikihow & 56.23 & 34.57 & 37.04 & 26.71 \\ %& 43.20 & 26.8 & 29.77 & 38.6 & 25.26 & 23.34 & 92.60 \\
        & \wikipedia & 47.94 & 21.21 & 27.77 & 16.06 \\ % & 33.78 & 8.07 & 1.01 & 38.78 & 33.45 & 34.5 & 57.18 \\
        & \outfox & 48.28 & 27.46 & 32.50 & 26.80 \\
        % & All & 78.57 & 37.80 & \textbf{47.86} & \textbf{35.53} \\ %& 45.16 & 18.85 & 46.54 & 56.60 & 48.02 & 26.48 & 93.36 \\
        % & \arxiv & 46.57 & 7.67 & \underline{11.55} & \underline{7.44} \\ %& 6.39 & 0.00 & 2.03 & 2.99 & 3.45 & 9.87 & 56.10 \\
        % & \peerread & 53.40 & 27.16 & 30.39 & 25.30 \\ % &17.27 & 0.58 & 6.15 & 60.72 & 5.64 & 33.37 & 89.01 \\
        % & \reddit & 61.96 & 28.23 & 33.07 & 26.53 \\ %& 50.18 & 15.22 & 10.78 & 26.11 & 29.06 & 12.92 & 87.18 \\
        % & \wikihow & 54.48 & 34.89 & 39.94 & 26.49 \\ %& 43.20 & 26.8 & 29.77 & 38.6 & 25.26 & 23.34 & 92.60 \\
        % & \wikipedia & 51.40 & 23.56 & 29.54 & 17.78 \\ % & 33.78 & 8.07 & 1.01 & 38.78 & 33.45 & 34.5 & 57.18 \\
        \midrule
        \multirow{6}{*}{NELA-SVM}
        & All & 64.50 & 23.91 & \textbf{30.54} & \textbf{22.76} \\ %& 37.64 & 0.36 & 18.63 & 30.52 & 48.00 & 14.75 & 93.72 \\
        & \arxiv & 47.35 & 11.53 & \underline{16.20} & \underline{10.94} \\ %& 13.64 & 0.07 & 12.88 & 10.73 & 19.00 & 10.81 & 66.62 \\
        & \peerread & 44.63 & 20.00 & 20.97 & 18.72 \\ %& 7.80 & 0.00 & 5.1 & 41.17 & 9.59 & 4.72 & 91.50 \\
        & \reddit & 42.77 & 24.27 & 27.87 & 20.72 \\ % & 49.00 & 0.46 & 8.69 & 8.32 & 48.28 & 13.96 & 91.43 \\
        & \wikihow & 48.33 & 25.81 & 25.51 & 21.83 \\ % & 48.68 & 0.20 & 6.96 & 13.88 & 16.80 & 16.65 & 96.59 \\
        & \wikipedia & 46.38 & 20.74 & 25.06 & 18.76 \\ % & 31.19 & 0.07 & 6.88 & 5.79 & 29.46 & 24.95 & 93.19 \\
        & \outfox & 35.05 & 17.44 & 18.48 & 19.18 \\
        % \multirow{6}{*}{NELA-SVM}
        % & All & 66.99 & 27.48 & \textbf{34.80} & \textbf{25.60} \\ %& 37.64 & 0.36 & 18.63 & 30.52 & 48.00 & 14.75 & 93.72 \\
        % & \arxiv & 53.46 & 12.97 & \underline{19.11} & \underline{11.71} \\ %& 13.64 & 0.07 & 12.88 & 10.73 & 19.00 & 10.81 & 66.62 \\
        % & \peerread & 48.53 & 24.76 & 22.84 & 23.31 \\ %& 7.80 & 0.00 & 5.1 & 41.17 & 9.59 & 4.72 & 91.50 \\
        % & \reddit & 45.94 & 28.09 & 31.45 & 24.20 \\ % & 49.00 & 0.46 & 8.69 & 8.32 & 48.28 & 13.96 & 91.43 \\
        % & \wikihow & 46.87 & 24.81 & 28.54 & 21.16 \\ % & 48.68 & 0.20 & 6.96 & 13.88 & 16.80 & 16.65 & 96.59 \\
        % & \wikipedia & 47.81 & 22.62 & 27.36 & 19.79 \\ % & 31.19 & 0.07 & 6.88 & 5.79 & 29.46 & 24.95 & 93.19 \\
        \bottomrule
    \end{tabular}
%    }
    \caption{\textbf{Task 2: multi-generator detection accuracy.} Classifiers are trained on the data of all domains except for the test domain (unseen). \textit{All} refers to the setting that randomly split train, validation and test sets, each has data of all domains. Given a detector, bold is the best and the underlined is the worst Acc and F1-macro.}
    \label{tab:subtaskB}
\end{table}

\begin{table*}
    \centering
    \small
%    \resizebox{\columnwidth}{!}{
    \begin{tabular}{l|l|c|c|c}
        \toprule
        \textbf{Detector} & \textbf{Train Data} & \textbf{Peerread} \textbf{\llamatwo-7B*} & \textbf{Peerread \chatgpt} & \textbf{All Test }\\
        \midrule
        \multirow{3}{*}{Longformer} 
         & All & 1.89 $\pm$ 0.79 & 4.36 $\pm$ 0.36 & 21.54 $\pm$ 0.25 \\
         & \chatgpt & 31.43 $\pm$ 6.15 & 4.55 $\pm$ 0.36 & 25.14 $\pm$ 0.93 \\
         & \llamatwo-7B* & 1.94 $\pm$ 0.07 &  51.379 $\pm$ 0.72 & 53.62 $\pm$ 1.60 \\
         \midrule
        \multirow{3}{*}{DeBERTa-v3}
         & All &     0.57 $\pm$ 0.23 & 2.63 $\pm$ 0.20 & 15.55 $\pm$ 2.60 \\
         & \chatgpt & 14.96 $\pm$ 2.19 & 2.53 $\pm$ 0.09 & 19.67 $\pm$ 1.05 \\
         & \llamatwo-7B* & 0.66 $\pm$ 0.12 & 24.59 $\pm$ 4.07 & 32.35 $\pm$ 0.78 \\         
        \bottomrule
    \end{tabular}
%    }
    \caption{\textbf{Task 3 MAE} for Longformer and Deberta-v3 under (1) cross-generator setting for PeerRead, and (2) unseen domains with multiple generators (\textit{All test}). Training data is \peerread using \llamatwo-7B* and \chatgpt.}
    \label{tab:subtaskC}
\end{table*}

\subsection{Multi-way Detection}

\paragraph{Experimental Setup}
% In this subtask, the detectors based on RoBERTa and XLM-R are trained to detect, which model was used to generate a text.
For task 2, the classifiers are trained to distinguish multiple generators (seven-label classification), including human, \davinci, \chatgpt, \gptfour, \cohere, \dolly, \bloomz. We measure the capabilities of detectors to generalize on a new domain that was not presented during training. The detectors are trained on the whole dataset except the domain used for testing. For this task, we calculate accuracy, macro F1-score, and the class-wise F1-scores.

We experiment with six baselines: fine-tuned RoBERTa and XLM-R, logistic regression (one-vs-rest) and SVM (one-vs-one) with GLTR features, and SVM (one-vs-one) with stylistic and NELA. % features.

% \paragraph{RoBERTa and XLM-R}
% TODO: ????

% \paragraph{LR vs. SVM with GLTR}
% Based on 14 GLTR features, we train logistic regression (one-vs-rest mode) and a SVM classifier (one-vs-one mode: an implementation of support vector machines) for multi-class classification.

% \paragraph{SVM with Stylistic or NELA Features}
% We used the same features as \secref{sec:subtaskA},
% With SVM, we aim to compare classification results using the same classifier but three different features: stylistic, NELA and GLTR.

\paragraph{Results and Analysis} 
% The results for task 2 are presented in \tabref{tab:subtaskB}.
We try to answer three questions: (1) given a detector, which domain is the hardest to predict; (2) which generator is the most challenging one to distinguish; and (3) which detector performs the best on task 2. 

As results in \tabref{tab:subtaskB}, Transformer-based detectors show poor performance in this task when facing with text from new domains not included in their training data. 
Taking RoBERTa as an example, low accuracy is presented when testing on any unseen domains, such as \arxiv and \wikipedia with accuracy of 36 and 52. Only when training over all domains and testing over all domains, accuracy reaches 97. 
The detectors exhibit significant challenges in generalizing to unfamiliar content, reflecting a notable weakness in their ability to handle previously unseen domains.
\textbf{The domain of \arxiv is the hardest} for most detectors to identify if they are trained without \arxiv examples.

% Experiments show that detectors perform well when they are trained and tested within the same domain. However the accuracy drops significantly when models are tested on unseen domains.
%Training on all and testing on all always obtain the best performance, especially by RoBERTa classifier. 
For most domains, \textbf{\davinci is the most challenging generator} to distinguish, followed by \chatgpt. \bloomz appears to be the easiest one, for all classifiers (see class-wise F1-scores in \appref{app:task2-results}). Even the worst detector NELA-SVM can identify \bloomz with >90\% F1-score. This may result from the fact that the generative distribution of \bloomz is significantly different from other generators, while the distribution between \davinci and \chatgpt are more similar to each other, making it hard to make distinctions.

\textbf{\roberta is the best} in this task followed by \xlmr.  Based on the \gltr features, SVM one-vs-one mode is better than logistic regression one-vs-rest mode. Under the same SVM, \gltr features are more helpful than stylistic and \nela.

\subsection{Boundary Identification}

\paragraph{Experimental Setup}

% The goal of Subtask C is to detect from which position the author of the text switches between human and machine.
% In our setting, the first segment is always written by human and then followed by machine-generated segments. 
% We need to predict the position of the boundary token.

% The training data for task 3 is based on the PeerRead domain and generations from \chatgpt and \llamatwo-7B. 
We perform evaluation for task 3 in two settings: (1) train on \peerread and test on \peerread (same or different generators); and (2) train on \peerread and test on a new domain with multiple generators (test column in \tabref{tab:dataset-C} with 11,123 examples). 

As a baseline, we use a sequence tagger based on Longformer~\citep{beltagy2020longformer} considering the long context (> 1,024 tokens) and DeBERTa-v3~\citep{he2021deberta},  where the position of the first word that is predicted as machine-generated is our prediction. The label was predicted at the token level: human-written words are tagged as 0 and machine-generated words are tagged as 1. We remap the tagged tokens back to the word level and calculate the MAE. We run experiments three times with different random seeds and calculate the mean and standard deviation. % of the evaluation scores. 

\paragraph{Results and Analysis}
In \tabref{tab:subtaskC}, when training on generations from \llamatwo \peerread and testing on \chatgpt \peerread and vice versa, even within the same domain, predicted boundary shows large gap with the gold label.
This reveals that it is much harder for boundary detection models to predict changing points for unseen generators.

Testing on an unseen domain \outfox with mixture of generators including \chatgpt, \gptfour and \llamatwo series, results in worse MAE, i.e., column \textit{all test}.
Taking the setting of \textit{All} as an example, MAE changes from <5 (column \peerread \llamatwo-7B* and \peerread \chatgpt) to MAE>21 and >15 (column \textit{all test}) respectively for Longformer and Deterta-v3.
Particularly for Longformer trained with \llamatwo data, MAE is greater than 53 when testing on \textit{all test}, demonstrating challenges on unseen domains and generators, consistent with findings in task 1 and 2.

% \section{Shared task Baseline}
% Using the same training dataset, we establish baselines for SemEval Task 8, and demonstrate results on development and test sets in this section for each subtask.

%% file: section/5_conclusion.tex
\section{Conclusion and Future Direction}

In this work, we extend M4 into a new benchmark M4GT-Bench for MGT detection, providing a wide coverage of nine languages, six domains, and nine state-of-the-art generators, including \gptfour and \llamatwo series.
We formulate three tasks, 
% While task 1 is a common binary MGT detection suggested in previous work, task 2 --- multi-class generator classification and 
with a novel formulation of detecting the changing point from human-written to machine-generated.
We observe that human readers are unable to consistently differentiate between generators and operate merely at the level of making random guesses.

For all tasks, 
% we provide strong baselines and analyze their performance in various evaluation settings. 
we find that the detector usually suffers a severe penalty in their performance when they are challenged with unseen domains or generators. Some generators such as \bloomz and domains like \arxiv appear harder than others. Overall, Transformer-based detectors such as \roberta and \xlmr are usually the best on average.

In future work, we plan to
% Based on the experiments and findings of our paper, we consider 
develop a demonstration system assisting human to detect MGT, especially in some high-stake domains such as clinical, financial and legal domains, and distinguish the authorship of different LLMs, promoting the protection of intelligence property.
In addition to text, we would extend to other modalities detection including speech, image and videos.
From the perspective of adversarial attacks such as paraphrasing, it is worthwhile to explore mixed texts with multiple changing points. % or vague boundaries.  

% as our main future direction. 
% As shown in the human evaluation part, it is difficult for human to differentiate accurately among generators even though there are some easily spottable patterns such as formatting, typos, and inconsistencies. However, recognizing that human focus on different features than detectors, it is interesting to development a system that can assist human to detect MGT, where, the system only give advice to human when human might make mistakes. 
% We remark that, since detectors might be biased toward non-native language speaker, it is important to involve human in this system in some high-stake area related to MGT. Another potential future direction is to study other adversarial settings such as paraphrase attack. It is also interesting to explore whether we can distinguish human written texts, machine generated texts and those mixed texts. Note that this is independent from task 3 and it is not necessarily easy since the mixed texts has features for both human and machine generated texts. 

% expand our M4 dataset continuously by introducing new LLM generators, by exploring different domains, by incorporating new languages, and by diversifying the range of tasks and prompts used. We believe that this is a good, practical way to keep the dataset up-to-date in response to the ongoing progress in LLMs. Our aim is to maintain a dataset that remains relevant as LLMs continue to evolve.

\section*{Acknowledgements}
We thank the anonymous reviewers for their insightful suggestions.

% \newpage
\section*{Limitations}
While we provide strong baselines and analyze their performance in various evaluation settings, certain limitations remain for future work. 
Firstly, the nature of supervised multi-way classification of different LLMs disables the distinction of unseen generators, and tends to fall into random guess once the distribution of training and test data differs.
Also, it is vulnerable to language style attack such as paraphrasing in different tones, back-translation and other textual adversarial attacks.
Methods based on watermark and few-shot in-content learning are more promising for this task.

Secondly, for all tasks, current black-box approaches may be less effective and may demonstrate poor generalization for unseen domains, generators, and languages, and this suggests the need to study more general methods to improve the detection and the potential misuse of LLMs. 

Thirdly, we simplify the task of boundary identification by assuming that the presented text is a mixture of first human-written segments and then machine-generated segments with only one changing point.
Practical scenarios are more complicated.
We should first detect whether it is a mixed text, and further identify all changing points.
% While we have conducted thorough experiments across various domains, languages, and generators, 
% One limitation is to include mixed texts that contains both human and machine generated texts as a new class to classify. 
% Specifically, in task 3, we assume the presence of texts with mixed authorship involving both humans and AI. However, it remains to explore whether these "mixed" texts can be easily identified before they are utilized for task 3. Although the above limitation is independent from our task formulation and is not a priority of our focus, we acknowledge the need for address this in the future work as a seperate task. 

\section*{Ethics and Broader Impact}
We discuss some potential ethical concerns about the present work.

\paragraph{Data Collection and Licenses}
We used pre-existing corpora: M4 and \outfox that have been publicly released and approved for research purposes, with clear dataset licenses.
Data licenses for multilingual dataset are listed in \tabref{tab:Task1-multilingual}. 

\paragraph{Security Implication}
The M4GT-Bench is intended for the development of a robust MGT detection system to identify and mitigate misuse, such as blocking the spread of automated misinformation campaigns, safeguarding individuals and institutions from potential financial losses. For domains of journalism, academia, and legal processes, the authenticity of information is paramount MGT detection ensures the integrity of content in these fields, thereby preserving trust. 
Also, users become more aware of LLMs capabilities, a growing skepticism towards digital content happens. Effective MGT detection can alleviate these concerns, ensuring users can trust LLM generations.

%% file: section/6_appendix.tex
% \todo[inline]{Add examples from new domains, generators, languages (only for those that were not part of M4 paper).}
\clearpage
\onecolumn
\section*{Appendix}
\appendix
\section{Task 3 Template Prompt}
\begin{figure*}[ht!]
	\centering
	\includegraphics[scale=0.63]{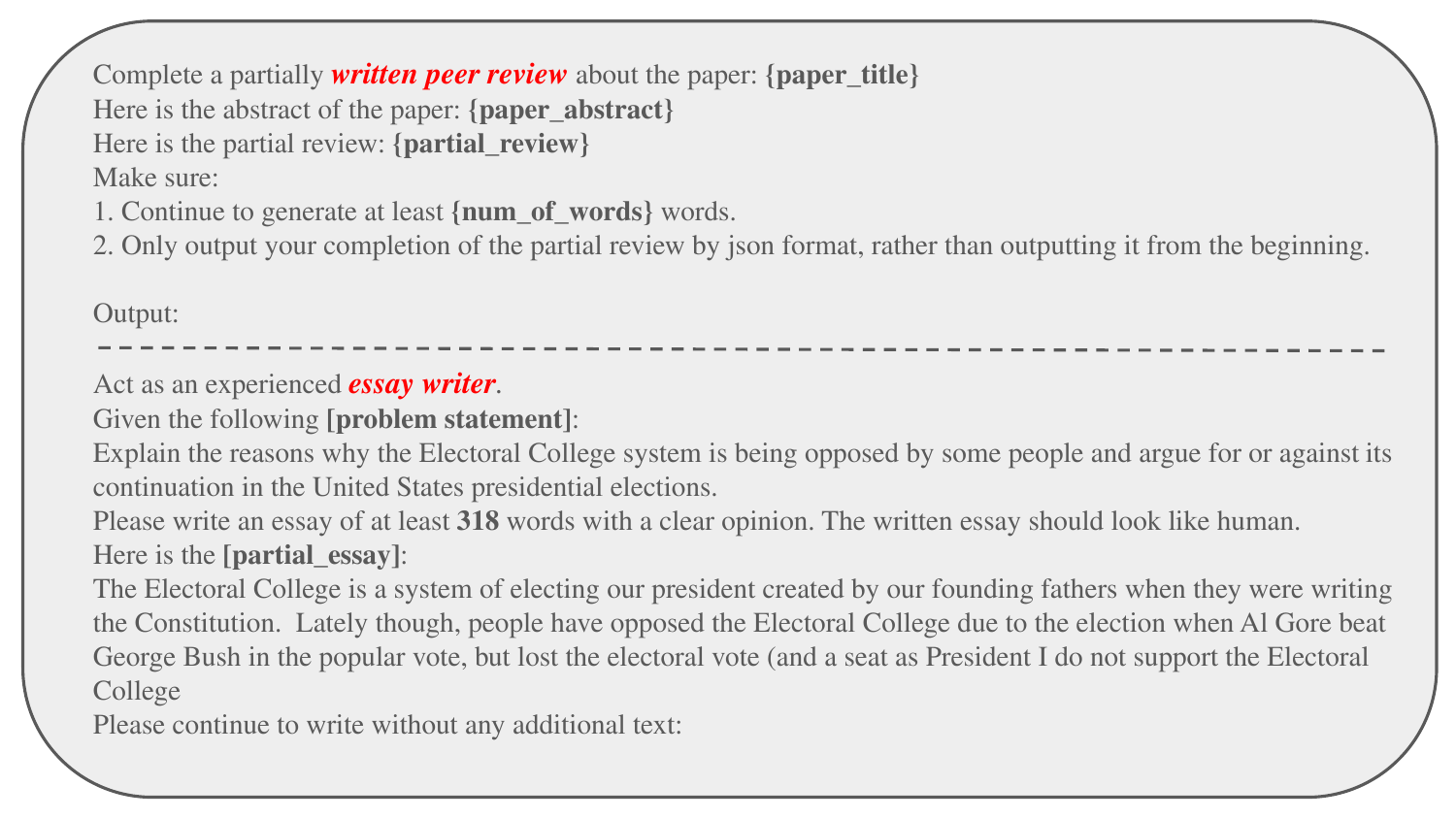}
	\caption{Task 3 prompt templates used to generate continuations of paper reviews and student essays.}
	\label{fig:subtaskC-prompt}
\end{figure*}

\figref{fig:subtaskC-prompt} provides template prompts used for task 3, instructing LLMs to continue to write peer reviews and student essays with human-written part as the context.

\section{Human Evaluation Results}
\label{app:humaneval}
\tabref{human-eval-full} presents the overall performance and generator-wise accuracy for human evaluation on task 2.

\begin{table*}[ht!]
    \centering
    \resizebox{\textwidth}{!}{
    \begin{tabular}{c|crrr|rrrrrrr}
        \toprule
      \multirow{2}{*}{Classifier}&  \multicolumn{4}{c|}{Overall Performance} & \multicolumn{7}{c}{Separate class F1-score} \\
       % & Prec & Recall & F1-macro\ & Acc &\chatgpt & Human & \cohere & \davinci & \bloomz &\dolly &\gptfour\\
        % Group 0 + Group 1& 22.09 & 19.73 & 21.10 & 20.27 & 22.22 & 25.53 & 10.0 & 32.0 & 0.0 & 10.0 & 42.11  \\
 % Group 2 + Group 3& 23.26 & 22.98 & 20.84 & 21.13 &  0.00 & 28.57 & 9.52 & 47.62 & 22.22 & 13.33 & 26.67  \\
 % All &              22.67 & 22.11 & 21.06 & 21.20 & 13.04 & 27.27 & 9.76 & 39.13 & 12.5 & 11.43 & 35.29 \\
       & Prec & Recall & F1-macro\ & Acc & Human & \davinci & \chatgpt & \gptfour & \cohere & \dolly & \bloomz \\
       \midrule
         Group 1&      \textbf{27.91} & \textbf{28.25} & \textbf{28.18} &\textbf{ 27.42} & 12.50 & 32.00 & 23.08 & 42.11 & --- & --- & --- \\
         Group 2&      16.28 & 09.22 & 12.91 & 10.45 & 32.26 & --- & --- & --- & 10.00 & 10.00 & 0.00 \\
         Group 3&      20.93 & 16.00 & 16.91 & 15.47 & 32.26 & --- & --- & --- & 9.52 & 13.33 & 22.22 \\
         Group 4&      25.58 & 25.83 & 25.25 & 24.82 & 25.00 & 47.62 & 0.00 & 26.67 & --- & --- & --- \\
         \midrule
         Group 1 + 2 & 22.09 & 19.73 & 21.10 & 20.27 & 25.53 & 32.00 & 22.22 & 42.11 & 10.00 & 10.00 & 0.00  \\
         Group 3 + 4 & 23.26 & 22.98 & 20.84 & 21.13 & 28.57 & 47.62 & 0.00 & 26.67 & 9.52 & 13.33 & 22.22 \\
         All &         22.67 & 22.11 & 21.06 & 21.20 & 27.27 & 39.13 & 13.04 & 35.29 & 9.76 & 11.43 & 12.50\\

    \bottomrule
    \end{tabular}
    }
   \caption{\textbf{Task 2 human evaluation:} overall performance and the F1-score for each separate class. Each group performs a four-class classification task. Results are even worse than random guess (25\%).}
    \label{human-eval-full}
\end{table*}

\clearpage
\section{Results of Multi-way Classification}
\label{app:task2-results}
\tabref{tab:subtaskB-full} provides the overall performance and class-wise (generator) F1-scores.
\begin{table*}[ht!]
    \centering
    \resizebox{\textwidth}{!}{
    \begin{tabular}{c|c|rrrr|rrrrrrr}
        \toprule
        \multirow{2}{*}{Detector}& Test & \multicolumn{4}{c|}{Overall Performance} & \multicolumn{7}{c}{Separate class F1-score} \\
        & Domain & Prec & Recall & F1-macro\ & Acc & Human & \davinci & \chatgpt & \gptfour & \cohere & \dolly & \bloomz \\
        \midrule
        \multirow{6}{*}{RoBERTa}
        & All & 96.96 & 97.01 & \textbf{96.94} & \textbf{97.00} & 96.77 & 96.53 & 97.75 & 96.22 & 97.96 & 94.00 & 99.34 \\
        & \arxiv & 55.72 & 36.55 & \underline{32.29} & \underline{36.55} & 32.70 & 1.61 & 33.88 & 47.59 & 8.48 & 38.55 & 63.24 \\
        & \peerread & 70.58 & 70.12 & 66.89 & 69.47 & 65.40 & 0.20 & 76.10 & 72.83 & 62.05 & 95.45 & 96.22 \\
        & \reddit & 77.21 & 74.49 & 71.66 & 74.49 & 84.47 & 75.44 & 44.94 & 35.51 & 81.84 & 82.41 & 96.97 \\
        & \wikihow & 72.62 & 70.56 & 68.85 & 68.36 & 38.91 & 41.52 & 64.73 & 90.41 & 62.32 & 85.09 & 98.97 \\
        & \wikipedia & 46.37 & 52.22 & 39.37 & 51.91 & 0.51 & 0.96 & 1.77 & 65.31 & 62.54 & 59.07 & 85.46 \\
        & \outfox & 71.71 & 65.04 & 66.40 & 78.25 & 72.63 & 57.80 & 87.00 & 0.00 & 73.46 & 74.28 & 99.63 \\
        \midrule
        \multirow{6}{*}{XLM-R}
        & All & 90.73 & 90.37 &\textbf{ 90.16} & \textbf{90.17} & 89.79 & 87.32 & 90.10 & 93.34 & 88.54 & 82.40 & 99.66 \\
        & \arxiv & 51.29 & 43.88 & 41.71 & 43.88 & 65.03 & 9.65 & 75.14 & 2.73 & 35.99 & 42.07 & 61.40 \\
        & \peerread & 53.68 & 52.14 & 46.10 & 50.91 & 20.69 & 9.09 & 66.43 & 34.33 & 34.77 & 63.92 & 93.50 \\
        & \reddit & 69.73 & 58.72 & 57.34 & 58.71 & 72.02 & 59.47 & 41.10 & 27.22 & 51.73 & 52.96 & 96.88 \\
        & \wikihow & 65.73 & 60.84 & 58.45 & 57.38 & 42.71 & 59.17 & 36.53 & 66.30 & 57.54 & 48.27 & 98.65 \\
        & \wikipedia & 60.04 & 42.55 & \underline{38.80} & \underline{41.95} & 5.50 & 45.40 & 38.07 & 29.29 & 40.25 & 50.44 & 62.63 \\
        & \outfox & 51.94 & 42.44 & 43.00 & 52.10 & 40.71 & 56.11 & 57.98 & 0.00 & 3.27 & 43.44 & 99.50 \\
        \midrule
        \multirow{6}{*}{GLTR-LR}
        % without outfox
        % & All & 42.9 & 46.11 & 43.07 & 45.84 & 66.98 & 13.15 & 37.06 & 30.17 & 50.30 & 33.83 & 70.00 \\
        % & \arxiv & 21.5 & 30.61 & \underline{23.19} & \underline{30.60} & 55.28 & 0.31 & 1.45 & 3.24 & 30.07 & 32.22 & 39.80 \\
        % & \peerread & 43.50 & 46.24 & 42.52 & 46.37 & 58.77 & 1.05 & 36.18 & 50.92 & 43.31 & 39.51 & 67.90 \\
        % & \reddit & 46.88 & 47.91 & \textbf{43.90} & \textbf{47.91} & 84.05 & 22.77 & 19.58 & 35.48 & 62.80 & 13.20 & 69.43 \\
        % & \wikihow & 43.87 & 38.78 & 38.28 & 37.23 & 57.49 & 14.69 & 28.41 & 11.35 & 54.79 & 18.79 & 82.44 \\
        % & \wikipedia & 41.91 & 35.66 & 35.54 & 34.08 & 67.54 & 11.78 & 14.10 & 20.12 & 32.22 & 33.90 & 69.14 \\
        & All & 42.36 & 43.96 & 40.32 & 45.06 & 65.13 & 11.98 & 40.22 & 14.39 & 44.88 & 33.0 & 72.65 \\
        & \arxiv & 26.24 & 34.45 & \underline{26.92} & \underline{34.45} & 58.38 & 0.06 & 17.78 & 0.34 & 37.59 & 35.84 & 38.47 \\
        & \peerread & 42.2 & 44.1 & 39.04 & 44.32 & 60.13 & 0.49 & 48.37 & 19.08 & 37.83 & 40.61 & 66.77 \\
        & \reddit & 45.54 & 46.28 & \textbf{41.7} & \textbf{46.28} & 85.06 & 24.08 & 22.44 & 19.77 & 55.74 & 14.6 & 70.23 \\
        & \wikihow & 41.86 & 39.39 & 38.24 & 38.74 & 59.14 & 13.92 & 32.06 & 1.03 & 57.84 & 19.52 & 84.16 \\
        & \wikipedia & 41.62 & 36.95 & 34.62 & 35.18 & 67.62 & 8.15 & 16.62 & 0.7 & 33.41 & 35.59 & 80.25 \\
        & \outfox & 29.68 & 32.38 & 29.18 & 38.78 & 49.79 & 12.24 & 37.5 & 0.0 & 1.71 & 17.51 & 85.49 \\

        \midrule
        \multirow{6}{*}{GLTR-SVM}
        % without outfox
        % & All & 53.19 & 53.47 & \textbf{52.50} & \textbf{53.75} & 70.15 & 32.16 & 53.37 & 42.57 & 53.57 & 41.41 & 74.28 \\
        % & \arxiv & 27.88 & 35.40 & \underline{29.25} & \underline{35.40} & 61.14 & 0.06 & 42.95 & 0.00 & 28.55 & 33.7 & 38.33 \\
        % & \peerread & 33.98 & 37.18 & 34.27 & 37.54 & 60.61 & 0.11 & 31.57 & 1.29 & 31.45 & 41.64 & 73.21 \\
        % & \reddit & 46.68 & 47.84 & 44.73 & 47.84 & 83.34 & 14.95 & 43.89 & 18.88 & 56.88 & 15.58 & 79.62 \\
        % & \wikihow & 43.88 & 42.36 & 40.52 & 37.84 & 66.13 & 22.54 & 1.15 & 23.16 & 64.09 & 23.65 & 82.93 \\
        % & \wikipedia & 43.62 & 37.96 & 35.69 & 37.18 & 53.55 & 17.18 & 40.28 & 43.48 & 32.03 & 36.89 & 26.45 \\
        & All & 52.81 & 48.42 & \textbf{47.24} & \textbf{50.39} & 69.35 & 21.39 & 49.47 & 31.21 & 46.84 & 36.04 & 76.35 \\
        & \arxiv & 22.57 & 32.29 & \underline{25.73} & 32.28 & 61.34 & 0.0 & 26.36 & 0.0 & 27.79 & 28.63 & 35.99 \\
        & \peerread & 34.1 & 39.1 & 34.81 & 39.4 & 60.64 & 0.15 & 41.71 & 0.17 & 30.74 & 36.98 & 73.26 \\
        & \reddit & 43.21 & 46.19 & 40.94 & 46.19 & 84.84 & 0.6 & 38.21 & 9.83 & 53.87 & 19.74 & 79.53 \\
        & \wikihow & 44.13 & 39.01 & 38.27 & 36.18 & 64.59 & 19.73 & 13.6 & 5.61 & 58.81 & 24.26 & 81.26 \\
        & \wikipedia & 34.86 & 30.95 & 26.72 & \underline{29.81} & 51.83 & 2.62 & 34.37 & 0.0 & 30.19 & 37.92 & 30.09 \\
        & \outfox & 26.93 & 28.29 & 26.45 & 28.58 & 51.25 & 27.8 & 2.84 & 0.0 & 0.54 & 15.71 & 86.99 \\

        \midrule
        \multirow{6}{*}{Stylistic-SVM}
        & All & 78.95 & 37.10 & \textbf{47.31} & \textbf{35.26} & 43.60 & 24.63 & 53.28 & 51.35 & 38.46 & 25.58 & 94.28 \\
        & \arxiv & 44.89 & 8.52 & \underline{12.71} & \underline{8.27} & 0.65 & 0.00 & 18.02 & 0.72 & 2.31 & 9.28 & 58.02 \\
        & \peerread & 50.72 & 21.96 & 25.43 & 20.44 & 11.23 & 0.50 & 7.78 & 28.56 & 3.57 & 37.50 & 88.85 \\
        & \reddit & 60.98 & 27.25 & 31.16 & 24.43 & 44.59 & 20.84 & 9.80 & 15.15 & 21.17 & 14.86 & 91.72 \\
        & \wikihow & 56.23 & 34.57 & 37.04 & 26.71 & 46.19 & 28.11 & 10.82 & 31.44 & 27.23 & 22.93 & 92.57 \\
        & \wikipedia & 47.94 & 21.21 & 27.77 & 16.06 & 33.62 & 8.46 & 1.77 & 31.31 & 22.78 & 30.43 & 66.03 \\
        & \outfox & 48.28 & 27.46 & 32.50 & 26.80 & 21.91 & 42.69 & 42.64 & 0.00 & 0.06 & 21.32 & 98.89 \\
        \midrule
        \multirow{6}{*}{NELA-SVM}
        & All & 64.50 & 23.91 & \textbf{30.54} & \textbf{34.11} & 0.18 & 20.58 & 19.97 & 28.86 & 15.26 & 94.82 \\
        & \arxiv & 47.35 & 11.53 & \underline{16.20} & \underline{10.94} & 4.34 & 0.00 & 18.55 & 3.80 & 7.32 & 11.67 & 67.74 \\
        & \peerread & 44.63 & 20.00 & 20.97 & 18.72 & 9.83 & 0.00 & 9.58 & 20.16 & 7.53 & 7.95 & 91.72 \\
        & \reddit & 42.77 & 24.27 & 27.87 & 20.72 & 43.73 & 0.20 & 7.84 & 2.62 & 31.48 & 17.52 & 91.67 \\
        & \wikihow & 48.33 & 25.81 & 25.51 & 21.83 & 47.41 & 0.13 & 1.59 & 2.52 & 15.15 & 15.11 & 96.63 \\
        & \wikipedia & 46.38 & 20.74 & 25.06 & 18.76 & 27.19 & 0.07 & 7.92 & 3.09 & 18.81 & 25.04 & 93.34 \\
        & \outfox & 35.05 & 17.44 & 18.48 & 19.18 & 2.00 & 0.20 & 24.84 & 0.00 & 0.00 & 2.66 & 99.68 \\
        \bottomrule
    \end{tabular}
    }
    \caption{\textbf{Task 2: multi-generator classification:} overall performance and the class-wise F1-scores. Classifier was trained on the data of all domains except for the test domain (unseen). \textit{All} refers to the setting that randomly split train, validation and test sets, each has data of all domains. In overall performance, given a detector, bold is the best and the underlined is the worst Acc and F1-macro.}
    \label{tab:subtaskB-full}
\end{table*}

\section{Examples of Task 1 and 2}
For English, we exhibit two examples from \gptfour-\arxiv and \gptfour-\outfox. 
For Arabic, we present two examples generated by \chatgpt and Jais-30B respectively based on news.
Afterwards, two examples generated by \chatgpt for German news and \wikipedia are demonstrated, followed by an Italian example produced by a fine-tuned \llamatwo-70B using Italian corpus, the model named \textsc{camoscio-70B}.

\clearpage
% \onecolumn
 \begin{center}
 \footnotesize
	\tablefirsthead{%
		\toprule
		   \textbf{Field} & \textbf{Content} \\
		\midrule}
	\tablehead{%
		\toprule
		   \textbf{Field} & \textbf{Content} \\
		\midrule}
	\tabletail{%
		\bottomrule
	}
	\tablelasttail{\bottomrule}
	\tablecaption{\textbf{Task 1 and 2 Examples} of English and other languages across different domains.}
	\begin{supertabular}{l | p{12cm}}
        % \midrule 
        Language &  \textbf{English} \\
        LLM Generator & \textsc{GPT-4} \\
        Domain &  \textit{\outfox} \\
        \midrule
        Problem Statement &  Explain the reasons why distractions while driving, specifically texting and driving, should be illegal and the consequences of not following this law. Use statistics and personal opinion to support your argument.\\
        \midrule
        Human & Distractions while driving could lead to deaths of other civilians. There are reasons why this topic became an actual law. The law was just passed last year that is you're on your phone while driving, that is an immediate ticket. Due to your criminal record (if it's bad) you can even do jail time if you feel you do not apply to this law. The cops of our cities are doing their job by stopping people who does not follow this law.  The main reason why this show not be condoned is, because most of these situations a citizen could get in a car accident. Approximately 1.6 million citizens die due to being on their cell phone while operating a vehicle. Nearly 390,000 people get real serious injuries due to texting and driving. One out of four of every car accident in the United States are due to texting and driving. In my opinion I believe texting and driving is the main reason we have so many deaths every year. texting and driving needs to be/ stay illegal. That way people will not do it so often, which could decrease the death rate each year. Texting and driving could possibly be the reason why our death rates increase over time. Especially in this day and age, because almost everything now is technology based  I believe that texting and driving is not safe for yourself or others, so technology should be the last thing you are on while behind and active vehicle. Texts, calls, emails all can wait until you are in a secure place (not behind the wheel) to respond, and also that way nobody could die or have an injury. \\ 
        \midrule
        Prompt & Consider the following problem statement: Explain the reasons why distractions while driving, specifically texting and driving, should be illegal and the consequences of not following this law. Use statistics and personal opinion to support your argument. Write a well-structured essay of approximately 325 words expressing your clear opinion on the issue. Your essay should provide a thorough analysis of the problem, explore potential implications, and propose viable solutions or recommendations. Be sure to consider different perspectives surrounding the problem and support your opinions with relevant evidence, examples, or logical reasoning. Additionally, emphasize the importance of addressing the issue and the potential impact of implementing your proposed solutions. Essay:\\
        \midrule
        Machine & Distractions while driving are a multi-faceted danger, with texting and driving sitting at the proverbial epicenter. The amalgamation of visual, manual, and cognitive distractions involved in this act makes it a leading hazard on the roads. This essay elucidates the imperative nature of legislation against texting and driving, buttressed by statistics and personal outlook, and explores the ramifications of non-compliance alongside potential resolutions. To initiate, consider the statistical landscape painting the grim reality. The National Highway Traffic Safety Administration (NHTSA) reports that in the United States alone, approximately 3,000 people lose their lives annually in crashes attributed to distracted driving. Texting is particularly pernicious; it takes the driver\u2019s eyes off the road for an average of five seconds, amounting to blind traversal of the length of a football field at 55 miles per hour. This nexus of inattention and control relinquishment is the breeding ground for tragedies. Moreover, the moral gravity of the situation cannot be overstated. When a driver engages in texting, they tacitly place their transient communication above the safety of fellow road users. This reflects a negligent societal value towards human life which by any humane standard is inexcusable. Hence, illegalizing such behavior manifests a collective ethos valuing life and responsibility. However, legal proscription must be complimented by earnest enforcement. Weak adherence to anti-texting laws, muddled by insubstantial sanctions, can subvert the intended deterrent effect. I assert that the consequence for distracted driving should encompass hefty fines, driving license suspensions, and, for repeat offenders or resulting in severe injury or death, significant custodial sentences. To mitigate the urge to reach for the phone, technological solutions like smartphone applications that disable texting while driving, and robust in-vehicle systems, need broader promotion. Educational campaigns fortifying public awareness about the consequences of distracted driving, seem pivotal to shifting attitudes, especially among impressionable young drivers. A symbiosis of legal strictures, technology, and education presents a tripartite solution to this problem. In conclusion, driving demands undivided attention, and texting fractures this focus. Legislating against such distractions is not only a proactive step towards reducing traffic-related fatalities and injuries, but it also reinforces a societal norm that prioritizes collective safety over momentary convenience. The significance of addressing this issue is underscored by the unbearable cost of human lives that could have been saved. In the grand tableau of road safety, cutting down the text during driving is not just a law; it is an ethical imperative that should carry the weight of universal endorsement.\\	
        & \\
        % \midrule 
        Language &  \textbf{English} \\
        LLM Generator & \textsc{GPT-4} \\
        Domain &  \textit{\arxiv} \\
        \midrule
        Title &  The Spitzer c2d Survey of Large, Nearby, Insterstellar Clouds. IX. The Serpens YSO Population As Observed With IRAC and MIPS\\
        \midrule
        Human & We discuss the results from the combined IRAC and MIPS c2d Spitzer Legacy observations of the Serpens star-forming region. In particular we present a set of criteria for isolating bona fide young stellar objects, YSO's, from the extensive background contamination by extra-galactic objects. We then discuss the properties of the resulting high confidence set of YSO's. We find 235 such objects in the 0.85 deg\^2 field that was covered with both IRAC and MIPS. An additional set of 51 lower confidence YSO's outside this area is identified from the MIPS data combined with 2MASS photometry. We describe two sets of results, color-color diagrams to compare our observed source properties with those of theoretical models for star/disk/envelope systems and our own modeling of the subset of our objects that appear to be star+disks. These objects exhibit a very wide range of disk properties, from many that can be fit with actively accreting disks to some with both passive disks and even possibly debris disks. We find that the luminosity function of YSO's in Serpens extends down to at least a few x .001 Lsun or lower for an assumed distance of 260 pc. The lower limit may be set by our inability to distinguish YSO's from extra-galactic sources more than by the lack of YSO's at very low luminosities. A spatial clustering analysis shows that the nominally less-evolved YSO's are more highly clustered than the later stages and that the background extra-galactic population can be fit by the same two-point correlation function as seen in other extra-galactic studies. We also present a table of matches between several previous infrared and X-ray studies of the Serpens YSO population and our Spitzer data set.\\ 
        \midrule
        Prompt & Write an abstract of an article from ArXiv given its title so that it is indistinguishable from a genuine human-written abstract. Write in the style that most abstracts from ArXiv are written. Title: The Spitzer c2d Survey of Large, Nearby, Insterstellar Clouds. IX. The Serpens YSO Population As Observed With IRAC and MIPS. Abstract: \\
        \midrule
        Machine & This paper presents the results of the Spitzer c2d survey of the Serpens star-forming region, focusing on the young stellar object (YSO) population as observed with the Infrared Array Camera (IRAC) and the Multiband Imaging Photometer for Spitzer (MIPS). The survey identified a significant number of YSO candidates, providing a comprehensive view of star formation in this region. The data obtained from IRAC and MIPS observations were used to construct color-color and color-magnitude diagrams, which were then utilized to classify the YSOs into different evolutionary stages. The spatial distribution of these objects was also analyzed, revealing a correlation with the dense molecular gas in the region. The results contribute to our understanding of the star formation process in large, nearby, interstellar clouds and provide a valuable resource for future studies.\\

        \midrule
        Language &  \textbf{Arabic} \\
        LLM Generator & \textsc{ChatGPT} \\
        Domain &  \textit{News} \\
        \midrule
        Title & {\small \begin{RLtext} الجيش التركي يعلن جاهزيته للعبور إلى ليبيا.. المعارك تتجدد بطرابلس وقوات حفتر تتراجع \end{RLtext}}\\
        \midrule
        Human &  {\small \begin{RLtext}
        تجددت المعارك اليوم الجمعة بمحيط العاصمة الليبية طرابلس، وأفادت التقارير\\بتراجع قوات اللواء المتقاعد خليفة حفتر ومقتل أحد أبرز قادتها، في حين أكد الجيش\\التركي جاهزيته للتوجه لليبيا فور تلقيه التعليمات. وقد أفاد مصدر عسكري من حكومة\\الوفاق بتجدد الاشتباكات المسلحة بأغلب محاور القتال جنوبي طرابلس. واستعادت قوات\\الوفاق مواقع خسرتها في وقت سابق، وأجبرت قوات حفتر على الانسحاب إلى خلف معسكر\\اليرموك بمحيط طرابلس. وذكرت مصادر من قوات الوفاق أن مقاتلي الخطوط الأمامية في\\قوات حفتر كانوا من مسلحي شركة ''فاغنر`` الروسية، الذين ينفذون عمليات قتالية ضد\\قوات حكومة الوفاق باستخدام أسلحة نوعية حديثة. وأكدت وسائل إعلام موالية لقوات\\حفتر أنها فقدت قائدا عسكريا بارزا يحمل رتبة عقيد ركن ويدعى فتحي محمد المسلاتي. \end{RLtext}}\\
        &  {\small \begin{RLtext}ولقي المسلاتي مصرعه اليوم الجمعة بقصف صاروخي في محور اليرموك جنوب العاصمة\\طرابلس. وكان المسلاتي يقود سلاح الدبابات والمدرعات باللواء 106 التابع لقوات\\حفتر. في السياق ذاته، استهدفت طائرات حفتر اليوم الجمعة تمركزات عسكرية بمحيط\\مصفاة الزاوية شمال غربي البلاد. وكانت المدينة قد تعرضت أمس الخميس لثلاث غارات\\جوية أسفرت إحداها عن مقتل مدنيين اثنين وجرح ثمانية آخرين نتيجة سقوط صواريخ على\\محلات عدة بالطريق الساحلي في الزاوية. واليوم الجمعة خرجت في المدينة مسيرة تندد\\بغارات قوات حفتر التي تستهدف المنشآت النفطية وتؤدي إلى سقوط ضحايا من المدنيين.\\وعبر المتظاهرون عن رفضهم للهجوم الذي تشنه قوات حفتر على العاصمة طرابلس، ونددوا\\بمواقف الدول الداعمة له. في انتظار الأوامرمن جانبها، أكدت مصادر عسكرية في\\أنقرة أن القوات المسلحة التركية مستعدة للتوجه إلى ليبيا والقيام بمهامها حال\\تلقيها التعليمات. جاء ذلك عقب مؤتمر صحفي للمتحدثة باسم وزارة الدفاع ناديدا\\شبنم أكتوب قدمت فيه تقييما لأنشطة الوزارة خلال العام 2019. وعن التوجه إلى\\ليبيا، قالت المصادر إن القوات المسلحة التركية مستعدة للقيام بكافة أنواع المهام\\داخل وخارج تركيا حال تلقيها التعليمات. وأمس الخميس، توقع الرئيس التركي رجب طيب\\أردوغان حصول حكومته على تفويض من البرلمان في الثامن من يناير/كانون الثاني\\المقبل من أجل إرسال جنود إلى ليبيا، تلبية لدعوة الحكومة الشرعية. وأكد أن بلاده\\ستقدم جميع أنواع الدعم لحكومة طرابلس في كفاحها ضد "الجنرال الانقلابي خليفة\\حفتر، المدعوم من دول أوروبية وعربية مختلفة". وفجر الخميس دخلت حيز التنفيذ\\مذكرة التفاهم الخاصة بالتعاون الأمني والعسكري بين تركيا وليبيا بعد نشرها في\\الجريدة الرسمية التركية. وفي وقت سابق، قال السفير الليبي في تركيا عبد الرزاق\\مختار عبد القادر إن بلاده تحتاج دعما عسكريا في الدفاع الجوي والتدريب الخاص،\\وذلك في ظل أنباء عن تقديم حكومة الوفاق طلبا رسميا إلى أنقرة للحصول على دعم\\عسكري بحري وبري وجوي. ومنذ 4 أبريل/نيسان الماضي، يشهد محيط طرابلس معارك مسلحة\\بين قوات حكومة الوفاق المعترف بها دوليا وقوات الجنرال المتقاعد خليفة حفتر.
        \end{RLtext}}\\
        % \midrule
        Prompt & {\small \begin{RLtext} 
        \LR{Given the following details for an Arabic news Article, generate the content in Arabic.\\Make sure the content is around 385 words.\\Title:\\ }الجيش التركي يعلن جاهزيته للعبور إلى ليبيا.. المعارك تتجدد بطرابلس وقوات حفتر تتراجع\\\LR{Description:\\} تجددت المعارك اليوم الجمعة بمحيط العاصمة الليبية طرابلس، وأفادت التقارير بتراجع قوات اللواء المتقاعد خليفة حفتر ومقتل أحد أبرز قادتها، بينما أكد الجيش التركي جاهزيته للتوجه لليبيا فور تلقيه التعليمات.\\\LR{Ref:\\} المصدر : الجزيرة + وكالات\\\\\\\LR{Content:\\}
        \end{RLtext}}\\
        \midrule
        Machine & 
        {\small \begin{RLtext}
        أفادت التقارير اليوم الجمعة بتجدد المعارك في منطقة طرابلس بليبيا، حيث يستمر الصراع المسلح بين القوات الموالية لحكومة الوفاق الوطني وقوات اللواء المتقاعد خليفة حفتر. وقد تم توثيق تراجع قوات حفتر، مع تقارير عن مقتل أحد قادتها البارزين خلال المعارك.\\\\في هذا السياق، أعلن الجيش التركي جاهزيته للتوجه إلى ليبيا، مشيرًا إلى أنه سيستجيب فورًا لأي تعليمات يتلقاها. ويأتي هذا الإعلان في ظل تصاعد التوتر في المنطقة وتأزم الأوضاع الأمنية في ليبيا.\\\\تأتي خطوة الجيش التركي في إطار مساعي الحكومة التركية لدعم حكومة الوفاق الوطني في ليبيا، التي تقاتل قوات حفتر وتسعى لحماية الاستقرار والأمن في البلاد. وقد تم تكثيف الاتصالات والتنسيق بين الجانبين لتعزيز العلاقات العسكرية والتنسيق السياسي.\\\\وتشعر الحكومة التركية بقلق شديد إزاء التطورات الأخيرة في ليبيا، حيث تشهد البلاد حربًا دموية مستعرة منذ أبريل 2019. وتأمل تركيا أن تلعب دورًا رئيسيًا في حل الأزمة وتحقيق الاستقرار في البلاد.\\\\يجب الإشارة إلى أن دخول الجيش التركي إلى ليبيا يستدعي موافقة البرلمان التركي، ولكن من المتوقع أن يتم تجاوز أي عوائق قانونية لتنفيذ هذه الخطوة. وقد صرح مسؤولون تركيون بأن بلادهم تتخذ كافة الإجراءات اللازمة لحماية مصالحها في ليبيا وضمان الأمن في البلاد.\\\\على صعيد متصل، تعتبر تركيا مستعدة للتعاون مع المجتمع الدولي من أجل تحقيق السلام والاستقرار في ليبيا. وتطالب الحكومة التركية بوقف العمليات العسكرية والعودة إلى الحوار السياسي، لتحقيق الانتقال الديمقراطي وتلبية تطلعات الشعب الليبي للسلام والحرية.
        \end{RLtext}}\\
        
       % \midrule
       % \midrule
        Language &  \textbf{Arabic} \\
        LLM Generator & \textsc{Jais} \\
        Domain &  \textit{News} \\
        \midrule
        Title & {\small \begin{RLtext}
        الجيش التركي يعلن جاهزيته للعبور إلى ليبيا.. المعارك تتجدد بطرابلس وقوات حفتر تتراجع
        \end{RLtext}}\\
        \midrule
        Human &  {\small \begin{RLtext}
        تجددت المعارك اليوم الجمعة بمحيط العاصمة الليبية طرابلس، وأفادت التقارير
بتراجع قوات اللواء المتقاعد خليفة حفتر ومقتل أحد أبرز قادتها، في حين أكد الجيش
التركي جاهزيته للتوجه لليبيا فور تلقيه التعليمات. وقد أفاد مصدر عسكري من حكومة
الوفاق بتجدد الاشتباكات المسلحة بأغلب محاور القتال جنوبي طرابلس. واستعادت قوات
الوفاق مواقع خسرتها في وقت سابق، وأجبرت قوات حفتر على الانسحاب إلى خلف معسكر
اليرموك بمحيط طرابلس. وذكرت مصادر من قوات الوفاق أن مقاتلي الخطوط الأمامية في
قوات حفتر كانوا من مسلحي شركة "فاغنر" الروسية، الذين ينفذون عمليات قتالية ضد
قوات حكومة الوفاق باستخدام أسلحة نوعية حديثة. وأكدت وسائل إعلام موالية لقوات
حفتر أنها فقدت قائدا عسكريا بارزا يحمل رتبة عقيد ركن ويدعى فتحي محمد المسلاتي.
ولقي المسلاتي مصرعه اليوم الجمعة بقصف صاروخي في محور اليرموك جنوب العاصمة
طرابلس. وكان المسلاتي يقود سلاح الدبابات والمدرعات باللواء 106 التابع لقوات
حفتر. في السياق ذاته، استهدفت طائرات حفتر اليوم الجمعة تمركزات عسكرية بمحيط
مصفاة الزاوية شمال غربي البلاد. وكانت المدينة قد تعرضت أمس الخميس لثلاث غارات
جوية أسفرت إحداها عن مقتل مدنيين اثنين وجرح ثمانية آخرين نتيجة سقوط صواريخ على
محلات عدة بالطريق الساحلي في الزاوية. واليوم الجمعة خرجت في المدينة مسيرة تندد
بغارات قوات حفتر التي تستهدف المنشآت النفطية وتؤدي إلى سقوط ضحايا من المدنيين.
وعبر المتظاهرون عن رفضهم للهجوم الذي تشنه قوات حفتر على العاصمة طرابلس، ونددوا
بمواقف الدول الداعمة له. في انتظار الأوامرمن جانبها، أكدت مصادر عسكرية في
أنقرة أن القوات المسلحة التركية مستعدة للتوجه إلى ليبيا والقيام بمهامها حال
تلقيها التعليمات. جاء ذلك عقب مؤتمر صحفي للمتحدثة باسم وزارة الدفاع ناديدا
شبنم أكتوب قدمت فيه تقييما لأنشطة الوزارة خلال العام 2019. وعن التوجه إلى
ليبيا، قالت المصادر إن القوات المسلحة التركية مستعدة للقيام بكافة أنواع المهام
داخل وخارج تركيا حال تلقيها التعليمات. وأمس الخميس، توقع الرئيس التركي رجب طيب
أردوغان حصول حكومته على تفويض من البرلمان في الثامن من يناير/كانون الثاني
المقبل من أجل إرسال جنود إلى ليبيا، تلبية لدعوة الحكومة الشرعية. وأكد أن بلاده
ستقدم جميع أنواع الدعم لحكومة طرابلس في كفاحها ضد "الجنرال الانقلابي خليفة
حفتر، المدعوم من دول أوروبية وعربية مختلفة". وفجر الخميس دخلت حيز التنفيذ
مذكرة التفاهم الخاصة بالتعاون الأمني والعسكري بين تركيا وليبيا بعد نشرها في
الجريدة الرسمية التركية. وفي وقت سابق، قال السفير الليبي في تركيا عبد الرزاق
مختار عبد القادر إن بلاده تحتاج دعما عسكريا في الدفاع الجوي والتدريب الخاص،
وذلك في ظل أنباء عن تقديم حكومة الوفاق طلبا رسميا إلى أنقرة للحصول على دعم
عسكري بحري وبري وجوي. ومنذ 4 أبريل/نيسان الماضي، يشهد محيط طرابلس معارك مسلحة
بين قوات حكومة الوفاق المعترف بها دوليا وقوات الجنرال المتقاعد خليفة حفتر.

        \end{RLtext}}\\
        
        % \midrule
        Prompt & {\small \begin{RLtext}
        بناءً على التفاصيل التالية لمقالة أخبار عربية، قم بإنشاء المحتوى باللغة العربية\\\\.تأكد من أن المحتوى يتراوح حول 385 كلمات.
\\\\العنوان: الجيش التركي يعلن جاهزيته للعبور إلى ليبيا.. المعارك تتجدد بطرابلس وقوات حفتر تتراجع
\\\\الوصف: تجددت المعارك اليوم الجمعة بمحيط العاصمة الليبية طرابلس، وأفادت التقارير بتراجع قوات اللواء المتقاعد خليفة حفتر ومقتل أحد أبرز قادتها، بينما أكد الجيش التركي جاهزيته للتوجه لليبيا فور تلقيه التعليمات.
المرجع: المصدر : الجزيرة + وكالات
المحتوى:
        \end{RLtext}}\\
        \midrule
        Machine & 
        {\small \begin{RLtext}
        الجيش التركي يعلن جاهزيته للعبور إلى ليبيا.. المعارك تتجدد بطرابلس وقوات حفتر تتراجع

أفادت التقارير الواردة من محيط العاصمة الليبية طرابلس بتجدد الاشتباكات اليوم الجمعة بين قوات حكومة الوفاق الوطني المعترف بها دوليا وقوات اللواء المتقاعد خليفة حفتر. كما أفادت التقارير بتراجع قوات حفتر ومقتل أحد أبرز قادتها.

وفي تطور آخر، أعلن الجيش التركي عن جاهزيته للعبور إلى ليبيا ونشر قواته فور تلقيه التعليمات من الحكومة التركية. وجاء هذا الإعلان بعد اجتماع للأمن القومي التركي يوم الخميس، برئاسة الرئيس رجب طيب أردوغان.

وكانت قوات حفتر قد حاولت في وقت سابق السيطرة على طرابلس، لكنها واجهت مقاومة شديدة من القوات الموالية للحكومة المعترف بها دوليا. وأسفرت المعارك الدائرة منذ أبريل / نيسان الماضي عن مقتل المئات وتشريد الآلاف.

من جهة أخرى، أكد الرئيس التركي رجب طيب أردوغان أن بلاده لن تظل صامتة إزاء قوات حفتر التي تدعمها الإمارات ومصر وروسيا. واتهم أردوغان هذه الدول بالتورط في سفك الدماء في ليبيا.

وفي سياق متصل، قال رئيس الحكومة الإيطالية جوزيبي كونتي إن بلاده تشعر بالقلق إزاء التطورات في ليبيا، مؤكدا أن ليبيا تحتاج إلى فعل وليس أقوال.

وتحاول الأمم المتحدة التوسط بين الأطراف المتناحرة في ليبيا لإنهاء القتال الدائر في البلاد. لكن المنظمة الدولية لم تتمكن من التوصل إلى تسوية سياسية حتى الآن.

وقد حذرت منظمات دولية من بينها الأمم المتحدة ومنظمة العفو الدولية من أن القتال في ليبيا قد وصل إلى مستويات جديدة من الرعب مع استهداف المدنيين والمدارس والمستشفيات.

ولا تزال الأوضاع في ليبيا تشهد تصعيدا على الأرض رغم الجهود الدولية لوقف إطلاق النار. ولا تزال قوات حفتر مستمرة في حصارها لطرابلس للشهر السادس على التوالي.

المراجع:
- المصدر : الجزيرة + وكالات

        \end{RLtext}}\\
        % & {\small \begin{RLtext}        
        % \end{RLtext}} \\
       
        % \midrule 
        Language &  \textbf{German} \\
        LLM Generator & \textsc{ChatGPT} \\
        Domain &  \textit{Wikipedia} \\
        \midrule
        Title &  Abwasserpumpe \\
        \midrule
        Human & Schmutzwasserpumpen sind in der Lage, auch grob verunreinigtes Wasser (aber auch andere fluide Medien) mit hohem Feststoffanteil wie Steinen, Geröll und Ähnlichem zu fördern. Sie arbeiten in der Regel nicht selbstansaugend, sondern sind als Tauchpumpe ausgelegt, bei denen ein Schaufelrad das Fluid fördert.
        Dafür werden besondere Laufradformen wie Freistromrad, Einkanalrad, Mehrkanalrad, Diagonalrad, Schraubenrad wie auch Propellerrad verwendet. Kleinere Pumpen haben meistens ein Freistromrad, da dies am wenigsten störanfällig ist und einen großen freien Durchgang für die Feststoffe bietet. Pumpen mit Freistromrädern haben jedoch schlechte Wirkungsgrade und erreichen auch keine großen Förderhöhen; deshalb wählt man bei größeren Maschinen meistens andere Laufradformen.
        Schmutzwasserpumpen sind auf die Fördermenge ausgelegt – typische Förderleistungen von 5.000 bis 15.000 Liter pro Stunde (etwa 3 bis 5 Liter je Sekunde) – und weniger auf den Druck, die Förderhöhe beträgt häufig nur einige Meter. Der Antrieb erfolgt in der Regel durch einen Elektromotor, je nach Anwendungsfall aber auch mittels eines Verbrennungsmotors erfolgen.
        Eingesetzt wird dieser Pumpentyp etwa im Tiefbau zum Auspumpen von Baugruben, im Hochwasserschutz und in der Katastrophenhilfe, zur Bewässerung in Landwirtschaft und Gartenbau oder in der Landschaftsarchitektur zur Befüllung von künstlichen Wasseranlagen. Kleinere Modelle werden auch im Heimbereich verwendet, um das Wasser für den Rasensprenger nicht dem „teuren“ Trinkwasser zu entnehmen, sondern aus nahen Fließgewässern oder Teichen (hierbei sind aber wasserrechtliche Vorschriften zu beachten).\\ 
        \midrule
        Prompt & This is a sample German Wikipedia summary section for the title "WBDJ":
        Der sozialistisch ausgerichtete Weltbund der Demokratischen Jugend (WBDJ) wurde am 10. November 1945 auf der Weltjugendkonferenz in London gegründet. Er ist von den Vereinten Nationen als internationale nichtstaatliche Jugendorganisation anerkannt.
        
        Dem WBDJ gehören über 150 Mitgliedsorganisationen (nach eigenen Angaben 2012 über 270) aus 109 Ländern an, die rund 30 Millionen Jugendliche vertreten. Diese Jugendorganisationen sind kommunistisch oder sozialistisch orientiert und/oder sie gehören  Befreiungsbewegungen an.
        
        Der WBDJ ist der Hauptträger der in unregelmäßigen Abständen seit 1947 stattfindenden Weltfestspiele der Jugend und Studenten.Der WBDJ war eine kommunistische Frontorganisation. In Folge des Zusammenbruchs des sozialistischen Lagers geriet der WBDJ 1990/91 in eine schwere Krise. Der künftige Charakter der Vereinigung wurde in Frage gestellt. Der WBDJ überstand jedoch diese Krise und stellte seine reguläre Arbeitsweise wieder her. 1997 wurden erstmals wieder Weltfestspiele auf Kuba durchgeführt. Höchstes Organ des WBDJ ist die Generalversammlung (bis 1957 Weltkongress), die zuletzt 2019 in Nikosia stattfand. Diese war mit 160 Vertretern von 93 Organisationen die größte Generalversammlung der letzten 30 Jahre.
        
        Using a similar formatting and style, write a German Wikipedia summary section for the title "Abwasserpumpe". The summary should have at least 250 words.\\
        \midrule
        Machine & Abwasserpumpe ist eine elektrische Pumpe zur Förderung von Abwasser und Schmutzwasser. Zusätzlich muss sie in der Lage sein, Feststoffe wie Haare, Papier oder Faserstoffe zu zerkleinern, um Verstopfung zu vermeiden. Man zählt die Abwasserpumpe zu den "Tauchmotorpumpen", da das gesamte Gerät unter Wasser getaucht wird. Das Pumpwerk (bestehend aus Saugkorb, Hydroscheider und Laufrad) wird dabei in einer Pumpstation eingesetzt. Der Antrieb des Laufrads geschieht durch einen Elektromotor. Die Pumpe trägt dazu bei, dass Abwasser und Schmutzwasser von der Einleitungsstelle zu den Kläranlagen oder sammelnden Kanalsystemen weitergeleitet werden. Insbesondere in Privathaushalten und Großküchen ist die Abwasserpumpe nützlich, da sie den Transport von Abwasser wesentlich erleichtert. Allerdings sollte man auf Handelsnamen achten: nicht alle Tauchmotorpumpen sind auch für den Einsatz in der Abwasserverarbeitung zugelassen.

        Eine häufig auftretende Schwierigkeit bei der Abwasserpumpe ist der sogenannte "Longpumpen". Hierbei kann der Abwasserstrom nicht mehr vollständig durch den Pumpenkopf geleitet werden, wodurch sich der Anteil sedimenter Stoffe erhöht. Darüber hinaus kann lang anhaltender Stress auf wenige Teile der Pumpe - etwa die Achse oder das Laufrad - schnell zu Abnutzungserscheinungen führen. Regelmäßige Wartung und Einhaltung von Grenzwerten für die Höhe des Abwasserdrucks oder die Größe der durchgeleiteten Partikel können diese Probleme vermindern.
        
        Die Qualität von Abwasserpumpen lässt sich anhand des maximalen zulässigen Drucks, der maximale Durchflussmenge, dem verwendeten Material oder auch in Hinblick auf Effizienz merklich voneinander unterscheiden. Es kann auch sinnvoll sein, beim Kauf einer Abwasserpumpe auf vorhandene Siegel und Zertifikate wie CE, VDE oder TÜV zu achten.\\	

        % \midrule
        % \midrule 
        Language &  \textbf{German} \\
        LLM Generator & \textsc{ChatGPT} \\
        Domain &  \textit{News} \\
        \midrule
        Title &  Journalisten und Publikum sägen an den Ästen, auf denen sie sitzen \\
        \midrule
        Human & Verlogen, unglaubwürdig, verwöhnt: So werden Journalisten heute taxiert. Sie werden alle in einen Topf geworfen, ihre Arbeit scheint wertlos, beliebig und ein Ärgernis. Das muss aufhören. Medienpublikum, Medienmanager, Medienforscher und Journalisten sägen derzeit selber an den Ästen, auf denen sie sitzen. Gewiss gibt es Journalismus, der über die Stränge schlägt oder abgehoben ist. Und es gibt zu viele Journalisten, die selbstgefällig sind, keine Fehler zugeben und sich diebisch freuen, wenn sie einen anderen in die Pfanne hauen können. Das sei nicht kleingeredet, aber ins Verhältnis gesetzt. Wir wissen, dass es Banker gibt, die in die eigene Tasche wirtschaften und zahllose Kunden ruinieren, Automobilclubs, die Ranglisten manipulierten, Unternehmer, die mit illegalen Abschalteinrichtungen Abgasnormen umgingen, Politiker, die politische Gegner ausspionierten, korrupte Fußballfunktionäre, quacksalbernde Ärzte, Winkeladvokaten. Und wir wissen das alles oft gerade, weil es eben neben schwarzen Schafen noch andere Journalisten gibt. Ende Februar wurde Spotlight mit einem Oscar ausgezeichnet: ein Film über eine Recherche, die die Missbrauchsfälle der katholischen Kirche in Boston enthüllte. Für diese Recherche gewann das Investigativteam des Boston Globe Jahre zuvor einen Pulitzerpreis. Doch nicht nur der Enthüllung von Missständen wegen müssen wir eine Lanze brechen für einen aufklärenden, verantwortungs- und werteorientierten Informationsjournalismus. Wir brauchen Medien, die uns Arenen öffnen, in denen wir um die besten Argumente ringen und diskutieren, wie wir künftig in unserer Gesellschaft leben wollen. Hören wir auf, an Ästen zu sägen, und fordern wir einen Journalismus ein, der sich bewusst seinem Publikum zuwendet, der Zweifel, Ängste und Sorgen ernst nimmt und auch konstruktiv aufzeigt, wie Probleme gelöst werden können.\\ 
        \midrule
        Prompt & Schreibe einen deutschen Nachrichtenartikel zu einem vorgegebenen Titel. Der Artikel muss mindestens 250 Wörter und höchstens 500 Wörter enthalten. Schreibe ausschließlich den Artikel und keine Anweisungen, Kommentare, Werbeanzeigen, Hashtags oder ähnliches. 

Titel: Journalisten und Publikum sägen an den Ästen, auf denen sie sitzen

Artikel: \\
        \midrule
        Machine & Heute leben wir in der Welt des Internets. Dies hat dazu beigetragen, dass jeder seinen Standpunkt darlegen und seine Meinung veröffentlichen kann. Aber haben wir dadurch eine Situation geschaffen, die zur Unabhängigkeit führt, oder haben wir uns in ein Gefängnis verwandelt, das von einer Gruppe von Menschen kontrolliert wird, die unsere Freiheit zensieren und unsere Perspektive beeinflussen?

        In den Tageszeitungen sind Journalisten in einer ähnlichen Situation gefangen. In dem Bestreben, die Meinungsfreiheit zu fördern und Nachrichten zu verbreiten, wird auch diese Gruppe von Menschen von der Unabhängigkeit des Internets gefangen genommen. Aber es gibt eine große Lücke zwischen der Zensur im Internet und der Zensur bei Tageszeitungen.

        Es ist wahr, dass Tageszeitungen dazu neigen, sich auf die Meinungen ihrer Leser zu konzentrieren, um einen zufriedenstellenden Inhalt bereitzustellen, der alle belohnt. Aber es ist auch wahr, dass diese Eigenarten der Tageszeitungen, als informelles Werkzeug für die politische Agenda genutzt werden kann. Werbewirtschaftliche und politisch motivierte Entscheidungen zwingen Journalisten dazu, die Wahrheit von dem zu unterscheiden, was derzeit trendig ist. 
        
        Die Leserschaft selbst, die mit Zeitungen und anderen Publikationen sozialisiert wurde, erwartet manchmal einen sehr spezifischen Nachrichteninhalt, der möglicherweise mehr mit ihrer eigenen ideologischen Perspektive als mit einer objektiven Wahrheit übereinstimmt. Eine solche Publikumsbindung kann dazu führen, dass Journalisten gezwungen werden, in eine bestimmte Richtung zu tendieren und sich zwangsläufig von der Wahrheit zu entfernen.
        
        Schließlich werden Journalisten dazu gezwungen, Artikel zu schreiben, die das Publikum möglicherweise nicht hören will. Sie schreiben über schwierige Themen wie Armut, Arbeitslosigkeit, Drogenmissbrauch, Hunger und Kriege. Aber einige Journalisten sind sich nicht sicher, wie sie über diese Probleme schreiben sollen, ohne das Publikum nicht zu verlieren, das dazu tendiert, einfache Geschichten mit einer positiveren Note zu bevorzugen. Dies kann dazu führen, dass Journalisten eine eingeschränkte Rolle in der Informationsbeschaffung übernehmen und das Publikum den verbleibenden Raum besetzt.
        
        Das Problem der Journalisten und Publikum, die an den Ästen sägen, auf denen sie sitzen, ist also ein beunruhigender Trend, der unsere Freiheit und Möglichkeiten beeinträchtigen kann. Wenn wir befreit und unabhängig bleiben wollen, benötigen wir mehr professionelle und mutige Journalisten, die bereit sind, eine Vielzahl von Lösungen für unser komplexes Leben zu präsentieren, unabhängig von den eigenen Interessen. Die Herausforderung besteht darin, das kritische Denken und die Unabhängigkeit der Medien zu fördern, um sicherzustellen, dass unsere Freiheit und unsere Weltsicht nicht von einer einflussreichen Gruppe von Menschen dominiert werden. Wir sollten das Bestreben der Journalisten in ihrem Bemühen um Unabhängigkeit unterstützen und uns dafür einsetzen, damit jeder die Freiheit hat, seine Meinung zu äußern und die Wahrheit zu suchen. \\	

        % \midrule 
        Language &  \textbf{Italian} \\
        LLM Generator & \textsc{camoscio-70B} \\
        Domain &  \textit{News} \\
        % \midrule
        % Question &  \\
        \midrule
        Human & {Francesco Schettino rilancerà il turismo delle Marche? Non c'è ancora nessuna ufficialità, ma già è scoppiata la bufera. Infatti, l\'ex comandante della Costa Concordia, secondo alcuni quotidiano locali, sarebbe pronto a diventare testimonial di Numana per il rilancio del turismo nella Riviera del Conero. Il sindaco però si è subito affrettato a precisare che si tratta di una idea malsana e fantasiosa. Sarà. Intanto però il legale di Schettino ha annunciato che il suo cliente avrà un colloquio di lavoro nei prossimi giorni. " Auspichiamo vivamente che questa buia fantasia non abbia fondamento alcuno: una persona divenuta famosa per un evento così negativo sicuramente non deve essere associata a una località che vive di turismo, bella e solare ", ha tuonato Carlo Neumann , presidente dell’Associazione Riviera del Conero. "Apprezzo lo sforzo di fantasia, ma bisogna partorire qualcosa di più interessante ", ha commentato il sindaco di Numana, Marzio Carletti (Pd). Insomma, ancora non si sa se Schettino metterà piede a Numana. Quello che è indubbio è che non ci sarà nessun inchino ad attenderlo} \\ 
        \midrule
        Prompt & \#\#\# Istruzione: Dato il testo 'Schettino testimonial per rilanciare il turismo nella Riviera del Conero?' scrivete un articolo di almeno 1000 parole in Italiano di cui quello è il titolo.
        \#\#\# Risposta: \\
        \midrule
        Machine & "La Riviera del Conero, situata sulla costa adriatica della regione Marche dell'Italia centrale, è una delle destinazioni turistiche più popolari d'Europa. È conosciuta per le sue spiagge di sabbia dorata e acque cristalline, così come per i suoi pittoreschi villaggi mediterranei che si snodano lungo la riva. Tuttavia, negli ultimi anni il numero di visitatori nella zona è diminuito significativamente a causa della crisi economica globale. Per cercare di invertire questa tendenza negativa, l'amministrazione locale ha deciso di rivolgersi al capitano Francesco Schettino, ex comandante della nave da crociera Costa Concordia naufragata nel 2012, per aiutare a promuovere nuovamente la Riviera del Conero come meta vacanziera ideale. Secondo gli amministratori locali, poiché il capitano Schettino è diventato un volto familiare in tutto il mondo grazie alla sua associazione con la tragedia della Costa Concordia, potrebbe essere uno strumento efficace per attirare attenzione sulla bellezza naturale ed esotica della Riviera del Conero"\\	
	\end{supertabular}
\end{center}

\clearpage
\section{Examples of Boundary Identification}
We demonstrate five examples below for boundary detection, including: \chatgpt-\peerread, \llamatwo-7B-\peerread, \gptfour-\outfox, \llamatwo-13B-\outfox and \llamatwo-70B-\outfox.

 \begin{center}
 \footnotesize
	\tablefirsthead{%
		\toprule
		   \textbf{Field} & \textbf{Content} \\
		\midrule}
	\tablehead{%
		\toprule
		   \textbf{Field} & \textbf{Content} \\
		\midrule}
	\tabletail{%
		\bottomrule
	}
	\tablelasttail{\bottomrule}
	\tablecaption{\textbf{Task 3 Examples} across different domains generated by \textsc{GPT} and \textsc{LLaMA2} series.}
	\begin{supertabular}{l | p{12cm}}
        % \midrule 
        Language &  \textbf{English} \\
        LLM Generator & \textsc{GPT-3.5-Turbo} \\
        Domain &  \textit{\peerread} \\
        \midrule
        Title & Learning to Compose Words into Sentences with Reinforcement Learning  \\
        Abstract & We use reinforcement learning to learn tree-structured neural networks for computing representations of natural language sentences. In contrast with prior work on tree-structured models, in which the trees are either provided as input or predicted using supervision from explicit treebank annotations, the tree structures in this work are optimized to improve performance on a downstream task. Experiments demonstrate the benefit of learning task-specific composition orders, outperforming both sequential encoders and recursive encoders based on treebank annotations. We analyze the induced trees and show that while they discover some linguistically intuitive structures (e.g., noun phrases, simple verb phrases), they are different than conventional English syntactic structures.  \\
         \midrule
        Prompt & Complete a partially written peer review about the paper: paper title Here is the abstract of the paper: paper abstract. Here is the partial review partial review. Make sure that
        1. the completion is of at least num of words words.
        2. that you only output the completion of the partial review and
        not write it from beginning.
        Output only the completion in the following json format.
        Make sure it can be parsed in python using json.loads:
        output format  \\
        \midrule
        Gold Boundary & 167 \\
        \midrule
        Human Segment & In this paper, the authors propose a new method to learn hierarchical representations of sentences, based on reinforcement learning. They propose to learn a neural shift-reduce parser, such that the induced tree structures lead to good performance on a downstream task. They use reinforcement learning (more specifically, the policy gradient method REINFORCE) to learn their model. The reward of the algorithm is the evaluation metric of the downstream task. The authors compare two settings, (1) no structure information is given (hence, the only supervision comes from the downstream task) and (2) actions from an external parser is used as supervision to train the policy network, in addition to the supervision from the downstream task. The proposed approach is evaluated on four tasks: sentiment analysis, semantic relatedness, textual entailment and sentence generation. I like the idea of learning tree representations of text which are useful for a downstream task. The paper is clear and well written. However, I am not convinced by the experimental results presented in the paper. \\ 
       
        \midrule
        Machine Segment & While the idea of learning tree structures for text representation is intriguing, the experimental results presented in the paper fail to adequately support the claims made by the authors. The comparison between the two settings, where no structure information is given and where actions from an external parser are used as supervision, is interesting. However, the authors do not provide enough evidence to demonstrate that the induced trees effectively improve performance on the downstream tasks. It is unclear whether the improvement can be attributed solely to the learned composition orders or if other factors contribute to the results. Additionally, the analysis of the induced trees falls short in providing a comprehensive understanding of their relationship to conventional English syntactic structures. The discovery of some linguistically intuitive structures, such as noun phrases and simple verb phrases, is promising. Nevertheless, a more in-depth analysis and comparison with established syntactic structures would strengthen the paper's argument. Overall, the paper is well-written and presents a novel approach, but it lacks sufficient empirical evidence and analysis to fully support its claims. \\	
        % \midrule

        % \midrule 
        Language &  \textbf{English} \\
        LLM Generator & \textsc{\llamatwo-7B} \\
        Domain &  \textit{\peerread} \\
        \midrule
        Title & Evaluation of Defensive Methods for DNNs against Multiple Adversarial Evasion Models \\
        Abstract & Due to deep cascades of nonlinear units, deep neural networks (DNNs) can automatically learn non-local generalization priors from data and have achieved high performance in various applications. However, such properties have also opened a door for adversaries to generate the so-called adversarial examples to fool DNNs. Specifically, adversaries can inject small perturbations to the input data and therefore decrease the performance of deep neural networks significantly. Even worse, these adversarial examples have the transferability to attack a black-box model based on finite queries without knowledge of the target model.  Therefore, we aim to empirically compare different defensive strategies against various adversary models and analyze the cross-model efficiency for these robust learners. We conclude that the adversarial retraining framework also has the transferability, which can defend adversarial examples without requiring prior knowledge of the adversary models. We compare the general adversarial retraining framework with the state-of-the-art robust deep neural networks, such as distillation, autoencoder stacked with classifier (AEC), and our improved version, IAEC, to evaluate their robustness as well as the vulnerability in terms of the distortion required to mislead the learner. Our experimental results show that the adversarial retraining framework can defend most of the adversarial examples notably and consistently without adding additional vulnerabilities or performance penalty to the original model. \\
         \midrule
        Prompt & Task:
Complete a partially-written peer review of the following paper.

Make sure that:

1. The completion is of at least 155 words.

2. You only complete the partial review and not write it from the beginning.

Title of the paper:
\{paper title\}

Abstract of the paper:
\{paper abstract\}

Review:
\{paper partial review\} \\
        \midrule
        Gold Boundary & 46 \\
        \midrule
        Human Segment & 
       I reviewed the manuscript as of December 6th.

The authors perform a systematic investigation of various retraining methods for making a classification network robust to adversarial examples. The authors achieve lower error rates using their RAD and IAEC methods perform better then previously introduced distillation methods for \\
        \midrule
        Machine Segment & 
        adversarial training. They demonstrate that their approach does not introduce any significant performance degradation compared to the baseline model. Overall, this work provides valuable insights into improving the robustness of deep learning models against adversarial attacks.\\	
        \midrule
        
        Language &  \textbf{English} \\
        LLM Generator & \textsc{GPT-4} \\
        Domain &  \textit{\outfox} \\
        \midrule
        Problem Statement & 
        Describe Luke's experience on the cattle boat trip and how it impacted his life. Include details about his background, the draft board's reaction to his trip, his work on the boat, and his overall perspective on the experience. Additionally, discuss how this experience influenced Luke's future interactions with international visitors. \\
        \midrule
        Prompt & Act as an experienced essay writer.
        
Given the following problem statement (essay topic), please write an essay of at least 152 words with a clear opinion. The written essay should look like human-written.

Partial essay:

Luke was at his part time job when his friend Don came in and said (do you want to come on a cattle boat trip with me.) Luke knew it was a opportunity of a life time and couldn't say no. Luke was \\
        
        \midrule
        Gold Boundary & 43 \\
        
        \midrule
        Human Segment & 
        Luke was at his part time job when his friend Don came in and said (do you want to come on a cattle boat trip with me.) Luke knew it was a opportunity of a life time and couldn't say no. Luke was 18 before he arived in Greece which meant he could be drafted for military service.
        
        When the draft board learned he was on a cattle baot trip they said that he was able to keep going on the cattle boat trip. The cattle boat trip was an unbeleivable opportunity for a small towns boy like hime.
        
        Luke helped out on his aunt Katie's farm whch prepared him for hard work but not for the dangers at sea. Luke also found a way to have fun on board even on return trips after the animals had been unloaded.
        
        When Luke got home he was very grateful that he was able to go on the trip. To the cowboys it was an adventure but to Luke it was a opportunity to open up a whole new world. And that awareness stayed with him, leading his family to host of intternational students and exchange visitors for many years. \\
        
        % \midrule
        Machine Segment & 
        Luke was at his part time job when his friend Don came in and said (do you want to come on a cattle boat trip with me.) Luke knew it was a opportunity of a life time and couldn't say no. Luke was a simple man from a small town in Kansas, with dreams bigger than the vast plains that surrounded him. The cattle boat trip was his first step towards realizing those dreams. The draft board, however, was not as enthusiastic about his trip. They saw it as a reckless adventure, a deviation from the norm. But for Luke, it was a chance to break free from the monotony of his life.

Working on the boat was far from easy. The long hours, the grueling physical labor, and the constant motion of the sea tested his endurance. But it also taught him resilience, patience, and the value of hard work. He learned to appreciate the simple joys of life - the beauty of a sunrise over the open sea, the camaraderie among the crew, and the satisfaction of a job well done.

The trip had a profound impact on Luke's life. It broadened his horizons and gave him a new perspective on life. He realized that the world was much bigger than his small town, and that there were countless cultures and ways of life to explore. This realization influenced his future interactions with international visitors. He welcomed them with open arms, eager to learn about their cultures and share his own experiences. He became a bridge between his small town and the rest of the world, fostering understanding and friendship among people from different backgrounds.

In conclusion, Luke's experience on the cattle boat trip was a turning point in his life. It challenged him, shaped him, and ultimately, enriched his life in ways he could never have imagined. It was a testament to the transformative power of travel and the endless possibilities it offers. \\	
        \midrule

        Language &  \textbf{English} \\
        LLM Generator & \textsc{\llamatwo-13B} \\
        Domain &  \textit{\outfox} \\
        \midrule
        Problem Statement &  Discuss the dangers of using cellular devices while driving and the impact it has on individuals and society. Include statistics and examples to support your argument.\\
        \midrule
        Gold Boundary & 38 \\
        \midrule
        Prompt & Act as an experienced essay writer.
        
        Given the following Problem statement (essay topic) complete the following Partial Essay writing at least 318 words with a clear opinion. The written essay should look like human-written (please start writing the essay without any additional text). 
        
        Problem statement (essay topic):
        
        Discuss the dangers of using cellular devices while driving and the impact it has on individuals and society. Include statistics and examples to support your argument. 
        
        Partial Essay: 
        
        Over the past 4 decades cell phones have evolved extremely. Nowadays people use their phones for just about any and everything. Things such as texting, calling, playing games, booking appointments, even checking latest posts on social media and much \\

        \midrule
        Human Segment & Over the past 4 decades cell phones have evolved extremely.
        
        Nowadays people use their phones for just about any and everything. Things such as texting, calling, playing games, booking appointments, even checking latest posts on social media and much more. Cell phones started to become a problem when people were just looking down at their phones while crossing the street. Something that could be fixed. Now, people are choosing to use their phones while driving which has become an issue and is very dangerous. Which puts not only the driver in danger put others as well.
        
        Each day people are dying or being injured from using their cellular devices while being behind the wheel. Each year 421,000 people in the United States are injured due to a driver being distracted by their device. It's more common for teens and young adults to text and drive (ages 18-24). 14\% of fatal crashes are involved from texting and driving in this age division. In 2017 the "Hands Free Law" had been issued. Which means drivers are allowed to use hand free devices only while being behind the wheel of a car and being completely free of a device. Although people may think the hands free law has prevented less accident it hasn't it only makes people have the urge to reach for their devices even more. 11 teenagers die every day due to texting and driving. Teenagers are 400\% more likely to get into an accident from texting and driving. AT\&T\'s Teen Driver Survey found that 97\% of teenagers think it's dangerous, while 43\% of them engage in the activity anyway.
        
        In conclusion I strongly disagree with texting and driving because it puts more than one person in danger. I feel as if you need to use your device while behind the wheel of a moving vehicle be mindful and pull to the side. Weather it\'s for making a quick call or sending a quick message even typing in a location into the GPS pull over because anything could happen in the blink of an eye.\\ 

        % \midrule
        Machine Segment & Over the past 4 decades cell phones have evolved extremely.
        
        Nowadays people use their phones for just about any and everything. Things such as texting, calling, playing games, booking appointments, even checking latest posts on social media and much more. However, one of the most dangerous things people do while using their phones is texting and driving. According to the National Highway Traffic Safety Administration (NHTSA), texting and driving has become one of the leading causes of car accidents in the United States. In 2018 alone, there were over 4,500 deaths and 140,000 injuries caused by motor vehicle crashes involving distracted drivers.
        
        Using cellular devices while driving poses a significant threat not only to the driver but also to passengers, pedestrians, and other drivers on the road. The act of texting and driving takes the driver's focus away from the road, causing them to be less aware of their surroundings. This lack of attention can lead to devastating consequences such as accidents, injuries, and even fatalities.
        
        Moreover, using cellular devices while driving can also have a negative impact on society as a whole. For instance, accidents caused by distracted driving can result in property damage, medical expenses, lost productivity, and even legal fees. These costs can be staggering and can have long-lasting effects on individuals, families, and communities. Additionally, the frequency of distracted driving incidents has led to increased insurance premiums for all drivers, which can be a significant financial burden for many people.
        
        Furthermore, the dangers of using cellular devices while driving are not limited to the driver alone. Pedestrians, cyclists, and other drivers are also at risk of being injured or killed by distracted drivers. In fact, according to the NHTSA, pedestrian fatalities have increased by 3.4\% from 2018 to 2019, with distracted driving being a major contributing factor. This highlights the need for stricter laws and regulations to prevent the use of cellular devices while driving.
        
        In conclusion, using cellular devices while driving poses a significant danger to individuals and society as a whole. The frequency of distracted driving incidents has led to increased accidents, injuries, and fatalities, as well as financial burdens for individuals, families, and communities. Therefore, it is crucial that we take measures to prevent the use of cellular devices while driving, such as implementing stricter laws and regulations, increasing public awareness campaigns, and investing in technology that can help reduce the temptation to use cellular devices while driving. By taking these steps, we can create safer roads for everyone and reduce the dangers associated with using cellular devices while driving.\\
        
        Language &  \textbf{English} \\
        LLM Generator & \textsc{\llamatwo-70B} \\
        Domain &  \textit{\outfox} \\
        \midrule
        Gold Boundary & 116 \\
        \midrule
        Problem Statement &  Explain the benefits of the Electoral College system and argue for its continuation in the United States presidential elections, citing its effectiveness over the past century and the role of electors in ensuring informed and responsible voting. \\
        \midrule
        Prompt &  Act as an experienced essay writer.
        
        Given the following Problem statement (essay topic) complete the following Partial Essay writing at least 318 words with a clear opinion. The written essay should look like human-written (please start writing the essay without any additional text).
        
        Problem statement (essay topic): 
        
        Explain the benefits of the Electoral College system and argue for its continuation in the United States presidential elections, citing its effectiveness over the past century and the role of electors in ensuring informed and responsible voting.
        
        Partial Essay:
        
        I would say that we keep the Electoral College because it has worked over the past century so why change it now. The people voting don't always know what they want in a president they might be undecided and just choose the best looking one I mean yeah some have their facts straight and know how the president they are voting is going to help the united states become a better country overall When people have to much freedom they don't know what to do and do wrong choices so i think choosing a elector is a good thing they have studied the president and have heard countless promises that he has done. I mean he\\
        % \midrule
        Human Segment & I would say that we keep the Electoral College because it has worked over the past century so why change it now. The people voting don\'t always know what they want in a president they might be undecided and just choose the best looking one I mean yeah some have their facts straight and know how the president they are voting is going to help the united states become a better country overall When people have to much freedom they don\'t know what to do and do wrong choices so i think choosing a elector is a good thing they have studied the president and have heard countless promises that he has done. I mean he or she most likley will vote for the candiate running for president you did because he or she is a elector he or she gets the people who also voted for the same person as he or she wants. I mean we people are still voting we still have the vote just that our vote goes to choosing a certain elector instead of just directly leaving it to the people to vote for a ceratin candiate running for president. By this the number is much smaller since each country is weighed and is worth certain points and decided by electors you choosed to trust to vote for the same president you did. You still are in control just that you choose somebody he/her is not likely to change his or her mind and vote for the exact person you want he or she to vote for. It stated in the article that "Each candidate running for president in your state has his or her own group of electors. The electors are generally chosen by the candiate\'s politcal party." I say that if the candidate political party chooses the electors then they must be good and the party must be sure the elector won\'t change his mind out of no where. This does make the candiate running for president try to convience the electors he is the one they need and that he will help our country be better. This also helps the candiate running for president know what states he should go to since some states they already know they lost It stated in the article that "A solid regional favorite, such as Romney was in the south,has no incentive to campaign heavily in those states, for he gains no electoral votes by increasing his plurality in states that he knows he will win."So this keeps the president from having to go to every state because its just no based of popular vote and want every individual person votes for. I think the Electoral college is great and well organized the process is good and the only thing is does is make the voting process more easier and ordered. I think we should keep it i mean it has had his troubles her and there but it has worked so good for such a long time. We don\'t want to change it and create choas since a lot of people are not use to change and don\'t take it quite well \\ 
        \midrule
        Machine Segment & I would say that we keep the Electoral College because it has worked over the past century so why change it now. The people voting don't always know what they want in a president they might be undecided and just choose the best looking one I mean yeah some have their facts straight and know how the president they are voting is going to help the united states become a better country overall When people have to much freedom they don't know what to do and do wrong choices so i think choosing a elector is a good thing they have studied the president and have heard countless promises that he has done. I mean he has to convince the elector that he is the right person for the job.
        
        The Electoral College system has been a cornerstone of American democracy for over two centuries, and it continues to serve the country well. Despite calls for its abolition, the system has proven its effectiveness time and time again, and it remains an essential component of the United States presidential elections. In this essay, I will argue for the continuation of the Electoral College system, citing its success over the past century and the critical role of electors in ensuring informed and responsible voting.
        
        First and foremost, the Electoral College system has a proven track record of effectively electing presidents who have gone on to serve the country with distinction. Over the past century, the system has produced some of the most iconic and influential leaders in American history, including Franklin D. Roosevelt, John F. Kennedy, and Barack Obama. These presidents have demonstrated exceptional leadership, vision, and commitment to the country, and their elections were made possible by the Electoral College system.
        
        Moreover, the Electoral College system ensures that the president is elected by a broad coalition of states, rather than just the most populous ones. This ensures that the president is accountable to voters across the country, rather than just those in heavily populated urban areas. For example, in the 2016 presidential election, Donald Trump won the presidency despite losing the popular vote to Hillary Clinton. However, he won a majority of the electoral votes, ensuring that he had a broad base of support across the country.
        
        Another significant advantage of the Electoral College system is the role of electors in ensuring informed and responsible voting. Electors are chosen by each state to cast votes for the president and vice president. They are typically party loyalists who have a deep understanding of the candidates and their policies. They are also responsible for ensuring that the president and vice president meet the constitutional requirements for office. This means that they must carefully consider the candidates' qualifications, character, and policy positions before casting their votes.
        \\ & 
        Furthermore, the Electoral College system provides a safeguard against the tyranny of the majority. In a direct popular vote system, the candidate who wins the most votes would automatically become the president, regardless of whether they have the support of a broad coalition of voters. This could lead to a situation where a candidate with extreme views or policies could win the presidency with a simple majority, even if they are opposed by a large majority of voters. The Electoral College system prevents this by requiring the president to win a majority of electoral votes, which ensures that the president has broad support across the country.
        
        Finally, the Electoral College system is an essential component of the federal system of government established by the Constitution. The system ensures that the president and vice president are elected by a coalition of states, rather than just the most populous ones. This ensures that the federal government remains accountable to the states and their citizens, rather than just the national population.
        
        In conclusion, the Electoral College system has been a cornerstone of American democracy for over two centuries, and it continues to serve the country well. It has a proven track record of effectively electing presidents who have gone on to serve the country with distinction, ensures that the president is accountable to voters across the country, provides a safeguard against the tyranny of the majority, and is an essential component of the federal system of government established by the Constitution. Therefore, I strongly argue for the continuation of the Electoral College system in the United States presidential elections.\\	
	\end{supertabular}
\end{center}

%% file: main.bbl
\begin{thebibliography}{44}
\expandafter\ifx\csname natexlab\endcsname\relax\def\natexlab#1{#1}\fi

\bibitem[{Bao et~al.(2023)Bao, Zhao, Teng, Yang, and Zhang}]{bao2023fast}
Guangsheng Bao, Yanbin Zhao, Zhiyang Teng, Linyi Yang, and Yue Zhang. 2023.
\newblock Fast-detectgpt: Efficient zero-shot detection of machine-generated text via conditional probability curvature.
\newblock \emph{arXiv preprint arXiv:2310.05130}.

\bibitem[{Beltagy et~al.(2020)Beltagy, Peters, and Cohan}]{beltagy2020longformer}
Iz~Beltagy, Matthew~E. Peters, and Arman Cohan. 2020.
\newblock \href {http://arxiv.org/abs/2004.05150} {Longformer: The long-document transformer}.
\newblock \emph{CoRR}, abs/2004.05150.

\bibitem[{Conneau et~al.(2019)Conneau, Khandelwal, Goyal, Chaudhary, Wenzek, Guzm{\'a}n, Grave, Ott, Zettlemoyer, and Stoyanov}]{conneau2019unsupervised}
Alexis Conneau, Kartikay Khandelwal, Naman Goyal, Vishrav Chaudhary, Guillaume Wenzek, Francisco Guzm{\'a}n, Edouard Grave, Myle Ott, Luke Zettlemoyer, and Veselin Stoyanov. 2019.
\newblock Unsupervised cross-lingual representation learning at scale.
\newblock \emph{arXiv preprint arXiv:1911.02116}.

\bibitem[{Crothers et~al.(2022)Crothers, Japkowicz, Viktor, and Branco}]{crothers2022adversarial}
Evan Crothers, Nathalie Japkowicz, Herna Viktor, and Paula Branco. 2022.
\newblock Adversarial robustness of neural-statistical features in detection of generative transformers.
\newblock In \emph{2022 International Joint Conference on Neural Networks (IJCNN)}, pages 1--8. IEEE.

\bibitem[{Dugan et~al.(2020)Dugan, Ippolito, Kirubarajan, and Callison-Burch}]{dugan-etal-2020-roft}
Liam Dugan, Daphne Ippolito, Arun Kirubarajan, and Chris Callison-Burch. 2020.
\newblock \href {https://doi.org/10.18653/v1/2020.emnlp-demos.25} {{R}o{FT}: A tool for evaluating human detection of machine-generated text}.
\newblock In \emph{Proceedings of the 2020 Conference on Empirical Methods in Natural Language Processing: System Demonstrations}, pages 189--196, Online. Association for Computational Linguistics.

\bibitem[{Dugan et~al.(2023)Dugan, Ippolito, Kirubarajan, Shi, and Callison{-}Burch}]{dugan2023realorfake}
Liam Dugan, Daphne Ippolito, Arun Kirubarajan, Sherry Shi, and Chris Callison{-}Burch. 2023.
\newblock \href {https://doi.org/10.1609/AAAI.V37I11.26501} {Real or fake text?: Investigating human ability to detect boundaries between human-written and machine-generated text}.
\newblock In \emph{Thirty-Seventh {AAAI} Conference on Artificial Intelligence, {AAAI} 2023, Thirty-Fifth Conference on Innovative Applications of Artificial Intelligence, {IAAI} 2023, Thirteenth Symposium on Educational Advances in Artificial Intelligence, {EAAI} 2023, Washington, DC, USA, February 7-14, 2023}, pages 12763--12771. {AAAI} Press.

\bibitem[{Gao et~al.(2024)Gao, Chen, Zhang, Huang, Wan, and Sun}]{gao2024llm}
Chujie Gao, Dongping Chen, Qihui Zhang, Yue Huang, Yao Wan, and Lichao Sun. 2024.
\newblock Llm-as-a-coauthor: The challenges of detecting llm-human mixcase.
\newblock \emph{arXiv preprint arXiv:2401.05952}.

\bibitem[{Gehrmann et~al.(2019{\natexlab{a}})Gehrmann, Strobelt, and Rush}]{gehrmann-etal-2019-gltr}
Sebastian Gehrmann, Hendrik Strobelt, and Alexander Rush. 2019{\natexlab{a}}.
\newblock \href {https://doi.org/10.18653/v1/P19-3019} {{GLTR}: Statistical detection and visualization of generated text}.
\newblock In \emph{Proceedings of the 57th Annual Meeting of the Association for Computational Linguistics: System Demonstrations}, pages 111--116, Florence, Italy. Association for Computational Linguistics.

\bibitem[{Gehrmann et~al.(2019{\natexlab{b}})Gehrmann, Strobelt, and Rush}]{gehrmann2019gltr}
Sebastian Gehrmann, Hendrik Strobelt, and Alexander~M Rush. 2019{\natexlab{b}}.
\newblock Gltr: Statistical detection and visualization of generated text.
\newblock \emph{arXiv preprint arXiv:1906.04043}.

\bibitem[{Gero et~al.(2022)Gero, Liu, and Chilton}]{gero2022sparks}
Katy~Ilonka Gero, Vivian Liu, and Lydia Chilton. 2022.
\newblock Sparks: Inspiration for science writing using language models.
\newblock In \emph{Designing interactive systems conference}, pages 1002--1019.

\bibitem[{Graham et~al.(2012)Graham, Haidt, Koleva, Motyl, Iyer, Wojcik, and Ditto}]{grahm2012hadit}
Jesse Graham, Jonathan Haidt, Sena Koleva, Matt Motyl, Ravi Iyer, Sean Wojcik, and Peter Ditto. 2012.
\newblock Moral foundations theory: The pragmatic validity of moral pluralism.
\newblock \emph{Advances in Experimental Social Psychology}, 47.

\bibitem[{Guo et~al.(2023)Guo, Zhang, Wang, Jiang, Nie, Ding, Yue, and Wu}]{guo2023close}
Biyang Guo, Xin Zhang, Ziyuan Wang, Minqi Jiang, Jinran Nie, Yuxuan Ding, Jianwei Yue, and Yupeng Wu. 2023.
\newblock \href {https://doi.org/10.48550/arXiv.2301.07597} {How close is chatgpt to human experts? comparison corpus, evaluation, and detection}.
\newblock \emph{CoRR}, abs/2301.07597.

\bibitem[{Hans et~al.(2024)Hans, Schwarzschild, Cherepanova, Kazemi, Saha, Goldblum, Geiping, and Goldstein}]{hans2024spotting}
Abhimanyu Hans, Avi Schwarzschild, Valeriia Cherepanova, Hamid Kazemi, Aniruddha Saha, Micah Goldblum, Jonas Geiping, and Tom Goldstein. 2024.
\newblock Spotting llms with binoculars: Zero-shot detection of machine-generated text.
\newblock \emph{arXiv preprint arXiv:2401.12070}.

\bibitem[{He et~al.(2021)He, Liu, Gao, and Chen}]{he2021deberta}
Pengcheng He, Xiaodong Liu, Jianfeng Gao, and Weizhu Chen. 2021.
\newblock \href {https://openreview.net/forum?id=XPZIaotutsD} {Deberta: decoding-enhanced bert with disentangled attention}.
\newblock In \emph{9th International Conference on Learning Representations, {ICLR} 2021, Virtual Event, Austria, May 3-7, 2021}. OpenReview.net.

\bibitem[{He et~al.(2023)He, Shen, Chen, Backes, and Zhang}]{he2023mgtbench}
Xinlei He, Xinyue Shen, Zeyuan Chen, Michael Backes, and Yang Zhang. 2023.
\newblock \href {https://doi.org/10.48550/ARXIV.2303.14822} {Mgtbench: Benchmarking machine-generated text detection}.
\newblock \emph{CoRR}, abs/2303.14822.

\bibitem[{Horne et~al.(2019)Horne, N{\o}rregaard, and Adali}]{horne2019robust}
Benjamin~D Horne, Jeppe N{\o}rregaard, and Sibel Adali. 2019.
\newblock Robust fake news detection over time and attack.
\newblock \emph{ACM Transactions on Intelligent Systems and Technology (TIST)}, 11(1):1--23.

\bibitem[{Hu et~al.(2023)Hu, Chen, and Ho}]{hu2023radar}
Xiaomeng Hu, Pin-Yu Chen, and Tsung-Yi Ho. 2023.
\newblock Radar: Robust ai-text detection via adversarial learning.
\newblock \emph{arXiv preprint arXiv:2307.03838}.

\bibitem[{Ippolito et~al.(2019)Ippolito, Duckworth, Callison-Burch, and Eck}]{ippolito2019automatic}
Daphne Ippolito, Daniel Duckworth, Chris Callison-Burch, and Douglas Eck. 2019.
\newblock Automatic detection of generated text is easiest when humans are fooled.
\newblock \emph{arXiv preprint arXiv:1911.00650}.

\bibitem[{Kirchenbauer et~al.(2023)Kirchenbauer, Geiping, Wen, Katz, Miers, and Goldstein}]{kirch2023watermark}
John Kirchenbauer, Jonas Geiping, Yuxin Wen, Jonathan Katz, Ian Miers, and Tom Goldstein. 2023.
\newblock \href {https://doi.org/10.48550/arXiv.2301.10226} {A watermark for large language models}.
\newblock \emph{CoRR}, abs/2301.10226.

\bibitem[{Koike et~al.(2023)Koike, Kaneko, and Okazaki}]{koike2023outfox}
Ryuto Koike, Masahiro Kaneko, and Naoaki Okazaki. 2023.
\newblock Outfox: Llm-generated essay detection through in-context learning with adversarially generated examples.
\newblock \emph{arXiv preprint arXiv:2307.11729}.

\bibitem[{Krishna et~al.(2023)Krishna, Song, Karpinska, Wieting, and Iyyer}]{krishna2023paraphrasing}
Kalpesh Krishna, Yixiao Song, Marzena Karpinska, John Wieting, and Mohit Iyyer. 2023.
\newblock Paraphrasing evades detectors of ai-generated text, but retrieval is an effective defense.
\newblock \emph{arXiv preprint arXiv:2303.13408}.

\bibitem[{Kumarage et~al.(2023)Kumarage, Garland, Bhattacharjee, Trapeznikov, Ruston, and Liu}]{kumarage2023stylometric}
Tharindu Kumarage, Joshua Garland, Amrita Bhattacharjee, Kirill Trapeznikov, Scott Ruston, and Huan Liu. 2023.
\newblock Stylometric detection of ai-generated text in twitter timelines.
\newblock \emph{arXiv preprint arXiv:2303.03697}.

\bibitem[{Li et~al.(2014)Li, Monaco, Chen, and Tappert}]{li2014authorship}
Jenny~S Li, John~V Monaco, Li-Chiou Chen, and Charles~C Tappert. 2014.
\newblock Authorship authentication using short messages from social networking sites.
\newblock In \emph{2014 IEEE 11th International Conference on e-Business Engineering}, pages 314--319. IEEE.

\bibitem[{Liu et~al.(2022)Liu, Zhang, Wang, Pu, Lan, and Shen}]{liu2022coco}
Xiaoming Liu, Zhaohan Zhang, Yichen Wang, Hang Pu, Yu~Lan, and Chao Shen. 2022.
\newblock Coco: Coherence-enhanced machine-generated text detection under data limitation with contrastive learning.
\newblock \emph{arXiv preprint arXiv:2212.10341}.

\bibitem[{Liu et~al.(2019)Liu, Ott, Goyal, Du, Joshi, Chen, Levy, Lewis, Zettlemoyer, and Stoyanov}]{liu2019roberta}
Yinhan Liu, Myle Ott, Naman Goyal, Jingfei Du, Mandar Joshi, Danqi Chen, Omer Levy, Mike Lewis, Luke Zettlemoyer, and Veselin Stoyanov. 2019.
\newblock Roberta: A robustly optimized bert pretraining approach.
\newblock \emph{arXiv preprint arXiv:1907.11692}.

\bibitem[{Macko et~al.(2024)Macko, Moro, Uchendu, Srba, Lucas, Yamashita, Tripto, Lee, Simko, and Bielikova}]{macko2024authorship}
Dominik Macko, Robert Moro, Adaku Uchendu, Ivan Srba, Jason~Samuel Lucas, Michiharu Yamashita, Nafis~Irtiza Tripto, Dongwon Lee, Jakub Simko, and Maria Bielikova. 2024.
\newblock Authorship obfuscation in multilingual machine-generated text detection.
\newblock \emph{arXiv preprint arXiv:2401.07867}.

\bibitem[{Mitchell et~al.(2023)Mitchell, Lee, Khazatsky, Manning, and Finn}]{mitchell2023detectgpt}
Eric Mitchell, Yoonho Lee, Alexander Khazatsky, Christopher~D. Manning, and Chelsea Finn. 2023.
\newblock \href {https://doi.org/10.48550/arXiv.2301.11305} {Detectgpt: Zero-shot machine-generated text detection using probability curvature}.
\newblock \emph{CoRR}, abs/2301.11305.

\bibitem[{Munir et~al.(2021)Munir, Batool, Shafiq, Srinivasan, and Zaffar}]{munir2021through}
Shaoor Munir, Brishna Batool, Zubair Shafiq, Padmini Srinivasan, and Fareed Zaffar. 2021.
\newblock Through the looking glass: Learning to attribute synthetic text generated by language models.
\newblock In \emph{Proceedings of the 16th Conference of the European Chapter of the Association for Computational Linguistics: Main Volume}, pages 1811--1822.

\bibitem[{Rivera~Soto et~al.(2024)Rivera~Soto, Koch, Khan, Chen, Bishop, and Andrews}]{rivera2024few}
Rafael Rivera~Soto, Kailin Koch, Aleem Khan, Barry Chen, Marcus Bishop, and Nicholas Andrews. 2024.
\newblock Few-shot detection of machine-generated text using style representations.
\newblock \emph{arXiv e-prints}, pages arXiv--2401.

\bibitem[{Shi et~al.(2023)Shi, Wang, Yin, Chen, Chang, and Hsieh}]{shi2023red}
Zhouxing Shi, Yihan Wang, Fan Yin, Xiangning Chen, Kai-Wei Chang, and Cho-Jui Hsieh. 2023.
\newblock Red teaming language model detectors with language models.
\newblock \emph{arXiv preprint arXiv:2305.19713}.

\bibitem[{Shu et~al.(2023)Shu, Luo, Hoskere, Zhu, Liu, Tong, Chen, and Meng}]{shu2023rewritelm}
Lei Shu, Liangchen Luo, Jayakumar Hoskere, Yun Zhu, Canoee Liu, Simon Tong, Jindong Chen, and Lei Meng. 2023.
\newblock Rewritelm: An instruction-tuned large language model for text rewriting.
\newblock \emph{arXiv preprint arXiv:2305.15685}.

\bibitem[{Solaiman et~al.(2019)Solaiman, Brundage, Clark, Askell, Herbert-Voss, Wu, Radford, Krueger, Kim, Kreps et~al.}]{solaiman2019release}
Irene Solaiman, Miles Brundage, Jack Clark, Amanda Askell, Ariel Herbert-Voss, Jeff Wu, Alec Radford, Gretchen Krueger, Jong~Wook Kim, Sarah Kreps, et~al. 2019.
\newblock Release strategies and the social impacts of language models.
\newblock \emph{arXiv preprint arXiv:1908.09203}.

\bibitem[{Su et~al.(2023)Su, Zhuo, Wang, and Nakov}]{su2023detectllm}
Jinyan Su, Terry~Yue Zhuo, Di~Wang, and Preslav Nakov. 2023.
\newblock Detectllm: Leveraging log rank information for zero-shot detection of machine-generated text.
\newblock \emph{arXiv preprint arXiv:2306.05540}.

\bibitem[{Uchendu et~al.(2020)Uchendu, Le, Shu, and Lee}]{uchendu2020authorship}
Adaku Uchendu, Thai Le, Kai Shu, and Dongwon Lee. 2020.
\newblock Authorship attribution for neural text generation.
\newblock In \emph{Proceedings of the 2020 conference on empirical methods in natural language processing (EMNLP)}, pages 8384--8395.

\bibitem[{Uchendu et~al.(2021)Uchendu, Ma, Le, Zhang, and Lee}]{uchendu2021turingbench}
Adaku Uchendu, Zeyu Ma, Thai Le, Rui Zhang, and Dongwon Lee. 2021.
\newblock Turingbench: A benchmark environment for turing test in the age of neural text generation.
\newblock \emph{arXiv preprint arXiv:2109.13296}.

\bibitem[{Venkatraman et~al.(2023)Venkatraman, Uchendu, and Lee}]{venkatraman2023gpt}
Saranya Venkatraman, Adaku Uchendu, and Dongwon Lee. 2023.
\newblock Gpt-who: An information density-based machine-generated text detector.
\newblock \emph{arXiv preprint arXiv:2310.06202}.

\bibitem[{Wang et~al.(2021)Wang, Xu, Wang, Gan, Cheng, Gao, Awadallah, and Li}]{wang2021adversarial}
Boxin Wang, Chejian Xu, Shuohang Wang, Zhe Gan, Yu~Cheng, Jianfeng Gao, Ahmed~Hassan Awadallah, and Bo~Li. 2021.
\newblock Adversarial glue: A multi-task benchmark for robustness evaluation of language models.
\newblock \emph{arXiv preprint arXiv:2111.02840}.

\bibitem[{Wang et~al.(2023)Wang, Mansurov, Ivanov, Su, Shelmanov, Tsvigun, Whitehouse, Afzal, Mahmoud, Aji et~al.}]{wang2023m4}
Yuxia Wang, Jonibek Mansurov, Petar Ivanov, Jinyan Su, Artem Shelmanov, Akim Tsvigun, Chenxi Whitehouse, Osama~Mohammed Afzal, Tarek Mahmoud, Alham~Fikri Aji, et~al. 2023.
\newblock M4: Multi-generator, multi-domain, and multi-lingual black-box machine-generated text detection.
\newblock \emph{arXiv preprint arXiv:2305.14902}.

\bibitem[{Xie et~al.(2023)Xie, Cohn, and Lau}]{xie2023next}
Zhuohan Xie, Trevor Cohn, and Jey~Han Lau. 2023.
\newblock The next chapter: A study of large language models in storytelling.
\newblock In \emph{Proceedings of the 16th International Natural Language Generation Conference}, pages 323--351.

\bibitem[{Xiong et~al.(2024)Xiong, Markchom, Zheng, Jung, Ojha, and Liang}]{xiong2024fine}
Feng Xiong, Thanet Markchom, Ziwei Zheng, Subin Jung, Varun Ojha, and Huizhi Liang. 2024.
\newblock Fine-tuning large language models for multigenerator, multidomain, and multilingual machine-generated text detection.
\newblock \emph{arXiv preprint arXiv:2401.12326}.

\bibitem[{Zellers et~al.(2019)Zellers, Holtzman, Rashkin, Bisk, Farhadi, Roesner, and Choi}]{zellers2019defending}
Rowan Zellers, Ari Holtzman, Hannah Rashkin, Yonatan Bisk, Ali Farhadi, Franziska Roesner, and Yejin Choi. 2019.
\newblock \href {https://proceedings.neurips.cc/paper/2019/hash/3e9f0fc9b2f89e043bc6233994dfcf76-Abstract.html} {Defending against neural fake news}.
\newblock In \emph{Advances in Neural Information Processing Systems 32: Annual Conference on Neural Information Processing Systems 2019, NeurIPS 2019, December 8-14, 2019, Vancouver, BC, Canada}, pages 9051--9062.

\bibitem[{Zhao et~al.(2023{\natexlab{a}})Zhao, Ananth, Li, and Wang}]{zhao2023provable}
Xuandong Zhao, Prabhanjan Ananth, Lei Li, and Yu-Xiang Wang. 2023{\natexlab{a}}.
\newblock Provable robust watermarking for ai-generated text.
\newblock \emph{arXiv preprint arXiv:2306.17439}.

\bibitem[{Zhao et~al.(2023{\natexlab{b}})Zhao, Wang, and Li}]{zhao2023protecting}
Xuandong Zhao, Yu{-}Xiang Wang, and Lei Li. 2023{\natexlab{b}}.
\newblock \href {https://doi.org/10.48550/arXiv.2302.03162} {Protecting language generation models via invisible watermarking}.
\newblock \emph{CoRR}, abs/2302.03162.

\bibitem[{Zhong et~al.(2020)Zhong, Tang, Xu, Wang, Duan, Zhou, Wang, and Yin}]{zhong2020neural}
Wanjun Zhong, Duyu Tang, Zenan Xu, Ruize Wang, Nan Duan, Ming Zhou, Jiahai Wang, and Jian Yin. 2020.
\newblock Neural deepfake detection with factual structure of text.
\newblock \emph{arXiv preprint arXiv:2010.07475}.

\end{thebibliography}
